\documentclass[10pt,twocolumn,letterpaper]{article}

\usepackage{cvpr}
\usepackage{times}
\usepackage{epsfig}
\usepackage{graphicx}
\usepackage{amsmath}
\usepackage{amssymb}
\usepackage{placeins}
\usepackage{booktabs}
\usepackage{xspace}

\usepackage{caption}
\usepackage{subcaption}
\usepackage{multirow}
\usepackage[toc,page]{appendix}

\usepackage[pagebackref=true,breaklinks=true,letterpaper=true,colorlinks,bookmarks=false]{hyperref}

\cvprfinalcopy

\newcommand\ganname{CIPS\xspace}

\begin{document}

\title{Image Generators with Conditionally-Independent Pixel Synthesis}

\author{I. Anokhin$^{1,2}$,  \;
K. Demochkin$^{1,2}$, \;
T. Khakhulin$^{1,2}$, \;
G. Sterkin$^1$, \;
V. Lempitsky$^{1,2}$, \;
D. Korzhenkov$^1$ \\
\\
$^1$Samsung AI Center, Moscow \\
$^2$Skolkovo Institute of Science and Technology, Moscow 
\vspace{-20pt}
}
\twocolumn[{%
    \renewcommand\twocolumn[1][]{#1}%
    \maketitle
    \centering
    \includegraphics[width=\textwidth]{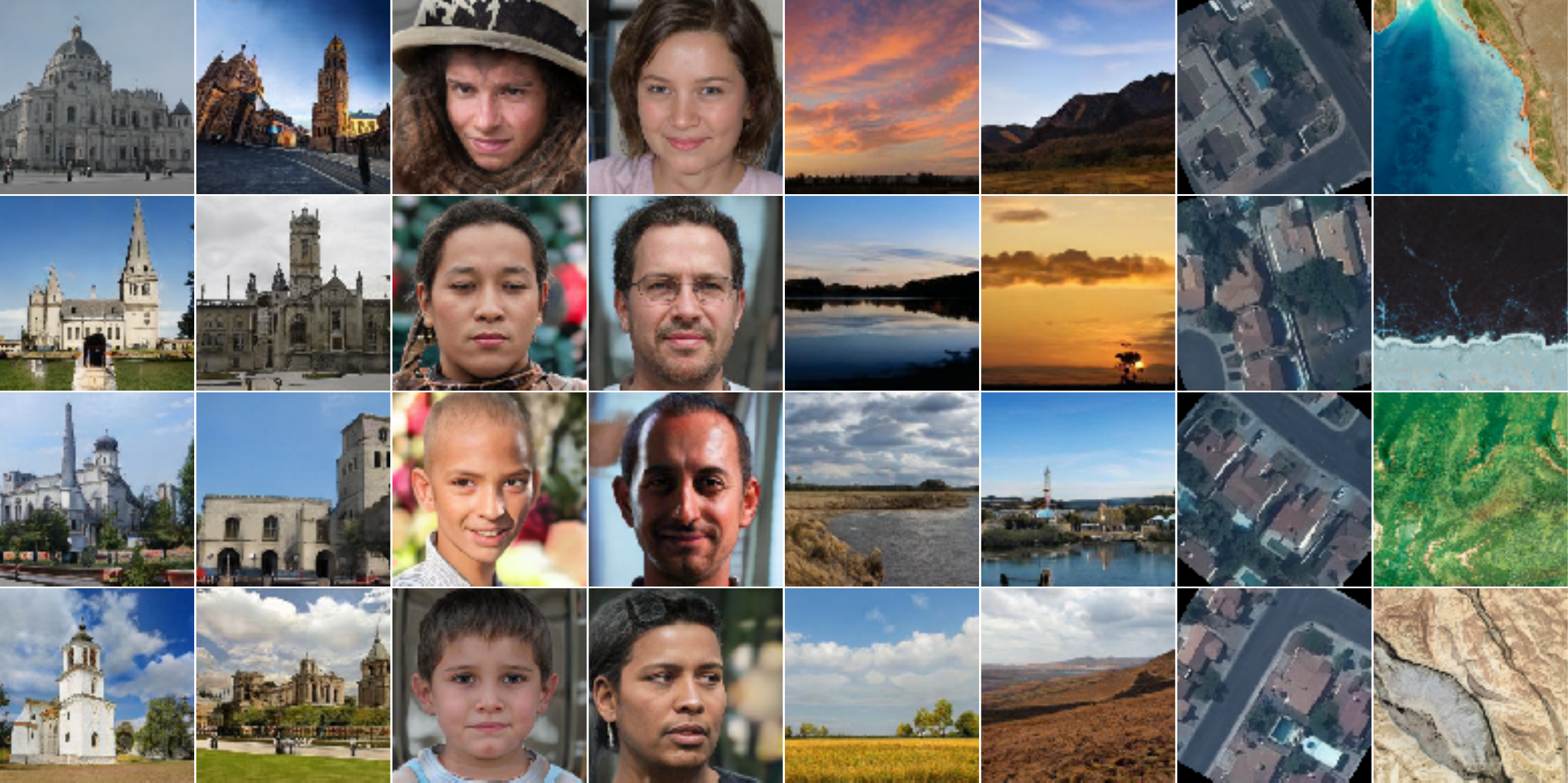}
    \captionof{figure}{Samples from our generators trained on several challenging datasets (LSUN Churches, FFHQ, Landscapes, Satellite-Buildings, Satellite-Landscapes) at resolution $256 \times 256$.
    The images are generated without spatial convolutions, upsampling, or self-attention operations.
    No interaction between pixels takes place during inference.
    }
    \label{fig:teaser}
    \vspace{.3cm}
    }]

\begin{abstract}
\vspace{-16pt}
Existing image generator networks rely heavily on spatial convolutions and, optionally, self-attention blocks in order to gradually synthesize images in a coarse-to-fine manner. Here, we present a new architecture for image generators, where the color value at each pixel is computed independently given the value of a random latent vector and the coordinate of that pixel. No spatial convolutions or similar operations that propagate information across pixels are involved during the synthesis. We analyze the modeling capabilities of such generators when trained in an adversarial fashion, and observe the new generators to achieve similar generation quality to state-of-the-art convolutional generators. We also investigate several interesting properties unique to the new architecture.
\end{abstract}

\section{Introduction}
\label{sec:introduction}

State-of-the-art in unconditional image generation is achieved using large-scale convolutional generators trained in an adversarial fashion~\cite{stylegan1,stylegan2,biggan}. While lots of nuances and ideas have contributed to the state-of-the-art recently, for many years since the introduction of DCGAN~\cite{dcgan} such generators are based around spatial convolutional layers, also occasionally using the spatial self-attention blocks~\cite{sagan}. Spatial convolutions are also invariably present in other popular generative architectures for images, including autoencoders~\cite{kingma2013auto}, autoregressive generators~\cite{van2016pixel}, or flow models~\cite{Dinh16,Kingma18}. Thus, it may seem that spatial convolutions (or at least spatial self-attention) is an unavoidable building block for state-of-the-art image generators.

Recently, a number of works have shown that individual images or collections of images of the same scene can be encoded/synthesized using rather different deep architectures (deep multi-layer perceptrons) of a special kind~\cite{nerf,siren}. Such architectures are not using spatial convolutions or spatial self-attention and yet are able to reproduce images rather well. They are, however, restricted to individual scenes. In this work, we investigate whether deep generators for unconditional image class synthesis can be built using similar architectural ideas, and, more importantly, whether the quality of such generators can be pushed to state-of-the-art.

Perhaps surprisingly, we come up with a positive answer (Fig.~\ref{fig:teaser}), at least for the medium image resolution (of $256\times{}256$). We have thus designed and trained deep generative architectures for diverse classes of images that achieve similar quality of generation to state-of-the-art convolutional generator StyleGANv2~\cite{stylegan2}, even surpassing this quality for some datasets.
Crucially, our generators are not using any form of spatial convolutions or spatial attention in their pathway. Instead, they use coordinate encodings of individual pixels, as well as sidewise multiplicative conditioning (weight modulation) on random vectors. Aside from such conditioning, the color of each pixel in our architecture is predicted independently (hence we call our image generator architecture \textit{Conditionally-Independent Pixel Synthesis} (CIPS) generators).

In addition to suggesting this class of image generators and comparing its quality with state-of-the-art convolutional generators, we also investigate the extra flexibility that is permitted by the independent processing of pixels. This includes easy extention of synthesis to non-trivial topologies (e.g.\ cylindrical panoramas), for which the extension of spatial convolutions is known to be non-trivial~\cite{cocogan,cohen2018spherical}.
Furthermore, the fact that pixels are synthesized independently within our generators, allows sequential synthesis for memory-constrained computing architectures. It enables our model to both improve the quality of photos and generate more pixel values in a specific areas of image (i.e.~to perform foveated synthesis).

\section{Related Work}
\label{sec:related}
Feeding pixel coordinates as an additional input to the neural network previously was successfully used in the widely known CoordConv technique~\cite{coordconv}  to introduce the spatial-relational bias. 
Recently, the same idea was employed by the COCO-GAN~\cite{cocogan} to generate images by parts or create "looped" images like spherical panoramas.
However, those models still used standard convolutions as the main synthesis operation. The synthesis process for neighboring pixels in such architectures is therefore not independent. 

To the best of our knowledge, the problem of regressing a given image from pixel coordinates with a perceptron (that calculates each pixel's value independently) started from creating compositional patterns with an evolutionary approach~\cite{cppn}.
Those patterns, appealing for digital artists, were also treated as kind of differentiable image parametrization~\cite{mordvintsev_xy2rgb}.
However, this approach was not capable of producing photorealistic hi-res outputs (e.g., see the demo~\cite{karpathy_regression}).

Some machine learning blogs reported experiments with GANs, where the generator was a perceptron that took a random vector and pixel coordinates as an input, and returned that pixel's value as an output~\cite{otoro1,otoro2}. The described model was successfully trained on MNIST, but has not been scaled to more complex image data.

Scene-representation networks~\cite{srn} and later the neural radiance fields (NeRF) networks~\cite{nerf} have demonstrated how 3D content of individual scenes can be encoded with surprising accuracy using deep perceptron networks. Following this realization, systems \cite{siren} and \cite{fourier_features} considered the usage of periodic activation functions and so-called Fourier features to encode the pixel (or voxel) coordinates, fed to the multi-layer perceptron. In particular, the ability to encode high-resolution individual images in this way was demonstrated. All these works however have not considered the task of learning image generators, which we address here.

The very recent (and independent) Generative Radiance Fields (GRAF) system~\cite{graf} showed promising results at embedding the NeRF generator into an image generator for 3D aware image synthesis. Results for such 3D aware synthesis (still limited in diversity and resolution) for certain  have been demonstrated. Here, we do not consider 3D-aware synthesis and instead investigate whether perceptron-based architectures can achieve high 2D image synthesis quality.

\section{Method}
\label{sec:method}

\begin{figure}[t]
    \centering
    \includegraphics[width=\columnwidth]{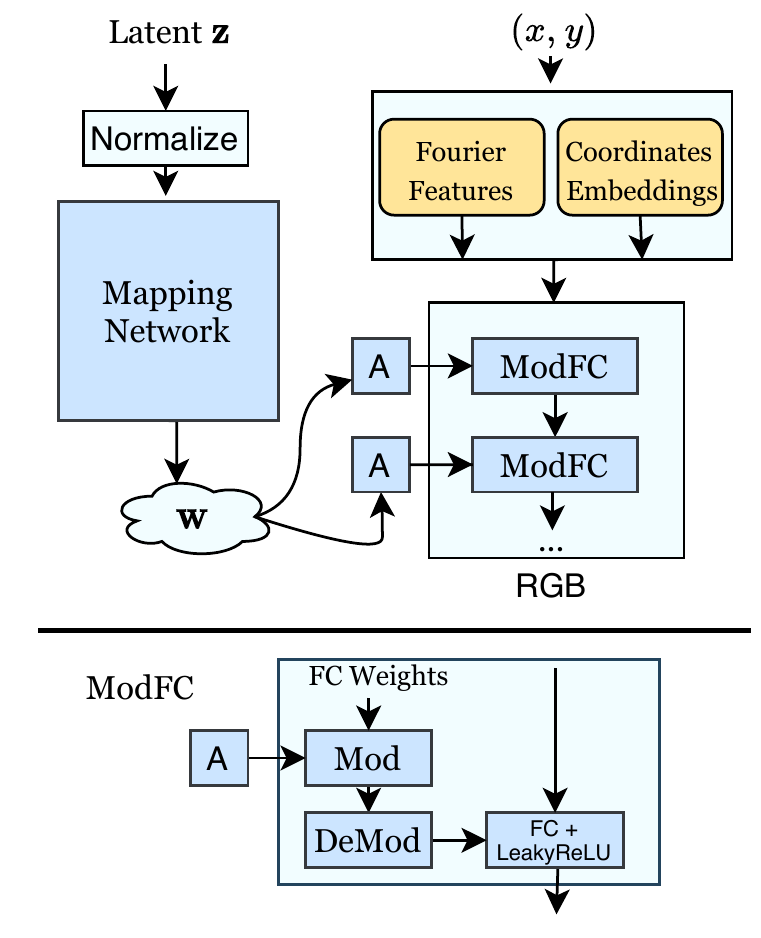}
    \caption{The Conditionally-Independent Pixel Synthesis (CIPS) generator architecture. 
        Top: the generation pipeline, in which the coordinates $\left(x, y\right)$ of each pixel are encoded (yellow) and processed by a fully-connected (FC) network with weights, modulated with a latent vector $\mathbf{w}$, shared for all pixels. The network returns the RGB value of that pixel. Bottom: The architecture of a modulated fully-connected layer (ModFC). Note: our default configuration also includes skip connections to the output (not shown here).
    }
    \label{fig:architecture}
\end{figure}

Our generator network synthesizes images of a fixed resolution $H \times W$ and has the multi-layer perceptron-type  architecture $G$ (see Fig.~\ref{fig:architecture}). In more detail, the synthesis of each pixel takes a random vector $\mathbf{z} \in \mathcal{Z}$ shared across all pixels, as well the pixel coordinates $\left(x, y\right) \in \left\{0\dots W-1\right\} \times \left\{0\dots H-1\right\}$ as input. It then returns the RGB value $\mathbf{c} \in \left[0, 1\right]^3$ of that pixel $G: \left(x, y, \mathbf{z}\right) \mapsto \mathbf{c}$.
Therefore, to compute the whole output image $I$, the generator $G$ is evaluated at at each pair $\left(x, y\right)$ of the coordinate grid, while keeping the random part $\mathbf{z}$ fixed:
\begin{align}
    I = \left\{ G\left(x, y; \mathbf{z}\right) \mid \left(x, y \right) \in \texttt{mgrid}\left(H, W\right) \right\},
    \label{eq:image_gen}
\end{align}
where 
\begin{equation*}
    \texttt{mgrid}\left(H, W\right) = \lbrace \left(x, y \right) \mid 0 \le x < W, \, 0 \le y < H  \rbrace
\end{equation*}
is a set of integer pixel coordinates.

Following~\cite{stylegan1}, a mapping network $M$ (also a perceptron) turns $\mathbf{z}$ into a \emph{style} vector $\mathbf{w} \in \mathcal{W}$, $M: \mathbf{z} \mapsto \mathbf{w}$, and all the stochasticity in the generating process comes from this style component.

We then follow the StyleGANv2~\cite{stylegan2} approach of injecting the style $\mathbf{w}$ into the process of generation via weight modulation.
To make the paper self-contained, we describe the procedure in brief here.

Any modulated fully-connected (ModFC) layer of our generator (see Fig.~\ref{fig:architecture}) can be written in the form $\mathbf{\psi} = \hat{B} \mathbf{\phi} + \mathbf{b}$, where $\mathbf{\phi} \in \mathbb{R}^n$ is an input, $\hat{B}$ is a learnable weight matrix $B \in \mathbb{R}^{m \times n}$ modulated with the style $\mathbf{w}$, $\mathbf{b} \in \mathbb{R}^m$ is a learnable bias, and $\mathbf{\psi} \in \mathbb{R}^m$ is an output.
The modulation takes place as follows: at first, the style vector  $\mathbf{w}$ is mapped with a small net (referred to as A in Fig.~\ref{fig:architecture}) to a scale vector $\mathbf{s} \in \mathbb{R}^n$
Then, the $\left(i, j\right)$-th entry of $\hat{B}$ is computed as
\begin{equation}
    \hat{B}_{ij} = \frac{s_j B_{ij}}{\sqrt{\epsilon + \sum\limits_{k=1}^n  \left(s_k B_{ik}\right)^2  }},
\end{equation}
where $\epsilon$ is a small constant.
After this linear mapping, a LeakyReLU function is applied to $\mathbf{\psi}$.

Finally, in our default configuration we add skip connections for every two layers from intermediate feature maps to RGB values and sum the contributions of RGB outputs corresponding to different layers. These skip connections naturally add values corresponding to the same pixel, and do not introduce interactions between pixels.

We note that the independence of the pixel generation process,  makes our model parallelizable at inference time and, additionally, provides flexibility in the latent space $z$. E.g., as we show below, in some modified variants of synthesis, each pixel can be computed with a different noise vector $z$, though gradual variation in $z$ is needed to achieve consistently looking images.

\subsection{Positional encoding}

The architecture described above needs an important modification in order to achieve the state-of-the-art synthesis quality.
Recently two slightly different versions of positional encoding for coordinate-based multi-layer perceptrons (MLP), producing images, were described in literature.
Firstly, SIREN~\cite{siren} proposed a perceptron with a principled weight initialization and sine as an activation function, used throughout all the layers.
Secondly, the Fourier features, introduced in \cite{fourier_features}, employed a periodic activation function in the very first layer only.
In our experiments, we apply a somewhat in-between scheme: the sine function is used to obtain Fourier embedding $e_{fo}$, while other layers use a standard LeakyReLU function: 
\begin{equation}
    e_{fo}\left(x, y\right) = \sin \left[B_{fo} \left(x', y'\right)^T \right],
\end{equation}
where $x' = \frac{2 x}{W-1}-1$ and $y' = \frac{2 y}{H-1} - 1$ are pixel coordinates, uniformly mapped to the range $\left[-1, 1\right]$ and the weight matrix $B_{fo} \in \mathbb{R}^{2 \times n}$ is learnable, like in SIREN paper.

However, only Fourier positional encoding usage turned out insufficient to produce plausible images. In particular, we have found out that the outputs of the synthesis tend to have multiple wave-like artifacts.
Therefore, we also train a separate vector $e_{co}^{\left(x, y\right)}$ for each spatial position and call them \textit{coordinate embeddings}. They represent $H \times W$ learnable vectors in total.
For comparison of these two embedding from the spectral point of view, see Sec.~\ref{sec:spectral_analysis}.
The full positional encoding $e\left(x, y\right)$ is a concatenation of  Fourier features and a coordinate embedding
\begin{equation}
    e\left(x, y\right) = \text{concat}\left[e_{fo}\left(x, y\right), e_{co}^{\left(x, y\right)}\right]
\end{equation}
and serve as an input for the next perceptron layer: $G\left(x, y;  \mathbf{z}\right) = G' \left( e\left(x, y\right);  M\left(\mathbf{z}\right) \right)$.

\section{Experiments}
\label{sec:experiments}

\subsection{Architecture details}

In our experiments, both Fourier features and coordinate embeddings had the dimension of 512.
The generator had 14 modulated fully-connected layers of width 512  
We use leaky ReLU activation with the slope $0.2$.
We implement our experiments on top of the public code\footnote{\url{https://github.com/rosinality/stylegan2-pytorch}} for StyleGANv2.
Our model is trained with a standard non-saturating logistic GAN loss with $R_1$ penalty~\cite{r1_penalty} applied to the discriminator $D$.
The discriminator has a residual architecture, described in \cite{stylegan2} (we have deliberately kept the discriminator architecture intact).
Networks were trained by Adam optimizer~\cite{adam} with learning rate $2 \times 10^{-3}$ and hyperparameters: $\beta_0=0$, $\beta_1=0.99$, $\epsilon=10^{-8}$.

\subsection{Evaluation}

We now evaluate CIPS generators and their variations on a range of datasets. For the sake of efficiency, most evaluations are restricted to $256 \times 256$ resolution. The following datasets were considered:

\begin{itemize}
    \item The \textit{Flickr Faces-HQ} (FFHQ)~\cite{stylegan1} dataset contains 70,000 high quality well-aligned, mostly near frontal human faces. This dataset is the most regular in terms of geometric alignement and the StyleGAN variants are known to perform very well in this setting.
    
    \item The \textit{LSUN Churches}~\cite{lsun} contains 126,000 outdoor photographs of churches of rather diverse architectural style. The dataset is regular, yet images all share upright orientation.
    
    \item The \textit{Landscapes} dataset contains 60,000 manually collected landscape photos from the Flickr website.
    
    \item The \textit{Satellite-Buildings}\footnote{\url{https://www.crowdai.org/challenges/mapping-challenge}} dataset contains 280,741 images of $300 \times 300$ pixels (which we crop to $256 \times 256$ resolution and randomly rotate). This dataset has large size, and is approximately aligned in terms of scale, yet lacks consistent orientation.  
    
    \item Finally, the \textit{Satellite-Landscapes}\footnote{\url{https://earthview.withgoogle.com/}} contains a smaller curated collection of 2,608 images of $512 \times 512$ resolution of satellite images depicting various impressive landscapes found on Google Earth (which we crop to $256 \times 256$ resolution). This is the most ``textural'' dataset, that lacks consistent scale or orientation. 
\end{itemize}

\begin{table}[t!]
\centering
\begin{tabular}{|l|c|c|c|} 
    \hline
    & StyleGANv2 & \ganname (ours) \\
    \hline
    FFHQ  & 3.83 & 4.38  \\ 
    LSUN Churches & 3.86 & 2.92 \\
    Landscapes & 2.36 & 3.61  \\
    
    Satellite-Buildings & 73.67 & 69.67  \\
    Satellite-Landscapes & 51.54 & 48.47  \\
    
    \hline
\end{tabular}
\caption{FID on multiple datasets at resolution of $256^2$ for CIPS-skips model. 
    Note that \ganname is of comparable quality with state-of-the-art StyleGANv2, and better on Churches. 
    The value for CIPS model on FFHQ differs from the one reported in Tab.~\ref{tab:modifications} as we trained this model for more time and with larger batch size. 
    }
\label{tab:fid}
\end{table}

\begin{table}[t]
\centering
\begin{tabular}{|l|c|c|} 
    \hline
    Model & Precision & Recall \\
    \hline
    StyleGANv2  & 0.609 & 0.513 \\ 
    CIPS  & 0.613 & 0.493 \\
    \hline
\end{tabular}
\caption{Precision \& Recall measured on FFHQ at $256^2$.
    The resulting quality of our model is better in terms of precision (corresponds to plausibility of images) and worse in recall (this points to the greater number of dropped modes).
    }
\label{tab:ffhq_prec_rec}
\end{table}

For evaluation, we relied on commonly used metrics for image generation: Frechet Inception Distance (FID)~\cite{fid} as well as more recently introduced generative Precision and Recall measures~\cite{precision_recall,improved_prec_rec}.

Our main evaluation is thus against the state-of-the-art StyleGANv2 generator~\cite{stylegan2}. We took FID value for LSUN Churches directly from the original paper~\cite{stylegan2} and trained StyleGANv2 on other datasets in authors' setup but without style-mixing and path regularization of generator -- as noted in the original paper, these changes do not influence the FID metric.  
The results of this key comparison are presented in Tab.~\ref{tab:fid} and \ref{tab:ffhq_prec_rec}. Neither of the two variants of the generator dominates the other, with StyleGANv2 achieving lower (better) FID score on FFHQ and Landscapes, while CIPS generator achieving lower score on LSUN Churches and both Satellite datasets.

\subsection{Ablations}
\label{sec:modifications}

\begin{table*}[h]
\centering
\begin{tabular}{|l|c|c|c|c|c|c|} 
    \hline
    CIPS                    & ``base'' & ``No embed (NE)'' & ``No Fourier'' & \textbf{Main} & ``Residual'' & Sine \\
    \hline
    Fourier Features          & + & + & -- & + & + & -- \\ 
    Coordinate Embeddings     & + & -- & + &+ & + & + \\ 
    Residual blocks           & -- & -- & -- &-- & + & -- \\
    Skip connections  & -- & -- & -- & + & -- & -- \\
    Sine Activation           & -- & -- & -- & -- & -- & + \\
    \hline
    FID  & 6.71 & 12.71 & 10.18 & \textbf{6.31} & 6.52 & 10.0 \\
    \hline
\end{tabular}
\caption{Effects of the modifications of CIPS generator on the FFHQ dataset in terms of Frechet Inception Distance (FID) score. Each column corresponds to a certain configuration, while rows correspond to present/missing features.
    The simultaneous usage of Fourier features and coordinate embeddings is necessary for a good FID score.
    Also, both residual connections and cumulative skip connections (default configuration) to the output outperform the plain multilayer perceptron.   
    }    
\label{tab:modifications}
\end{table*}

\begin{figure}[t]
    \centering
    \includegraphics[width=\columnwidth]{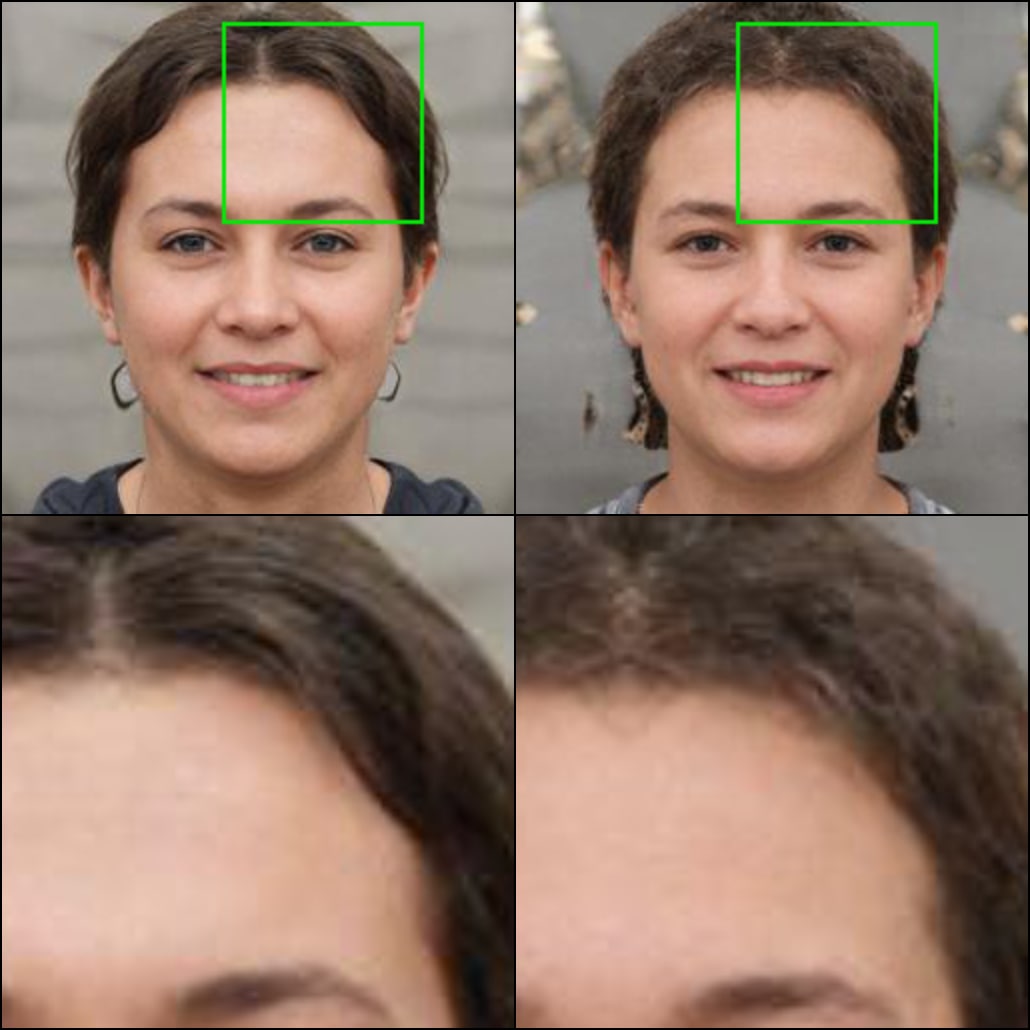}
    \caption{Image corresponding to the mean style vector in the space of $\mathcal{W}$ for \ganname (left) and  \ganname-NE (no embeddings) generators (right). Left image has more plausible details like hair which confirms the results in Tab.~\ref{tab:modifications}. 
    }
    \label{fig:cips-vs-cips-wo-emb}
\end{figure}

We then evaluate the importance of different parts of our model by its ablation on the FFHQ dataset (Tab.~\ref{tab:modifications}). 
We thus consider removing Fourier features, coordinate embeddings (config referred to as \emph{\ganname-NE}) and replace LeakyReLU activation with sine function in all layers. 
We also compare the variants with residual connections (we follow StyleGANv2~\cite{stylegan2} implementation adjusting variance of residual blocks with the division by $\sqrt{2}$) with our main choice of cumulative projections to RGB.
Additionally, the ``base'' configuration without skip connections and residual connections is considered.
In this comparison, all models were trained for 300K iterations with batch size of 16.

As the results show, coordinate embeddings, residual blocks and cumulative projection to RGB significantly improve the quality of the model. 
The  removal of coordinate embeddings  most severely worsens the FID value, and affects the quality of generated images (Fig.~\ref{fig:cips-vs-cips-wo-emb}). We further investigate the importance of coordinate embeddings for the CIPS model below. 

\subsection{Influence of positional encodings}

\begin{figure}[ht]
    \centering
    \setlength{\tabcolsep}{2pt}
    \begin{tabular}{cc}
        \includegraphics[width=0.49\columnwidth]{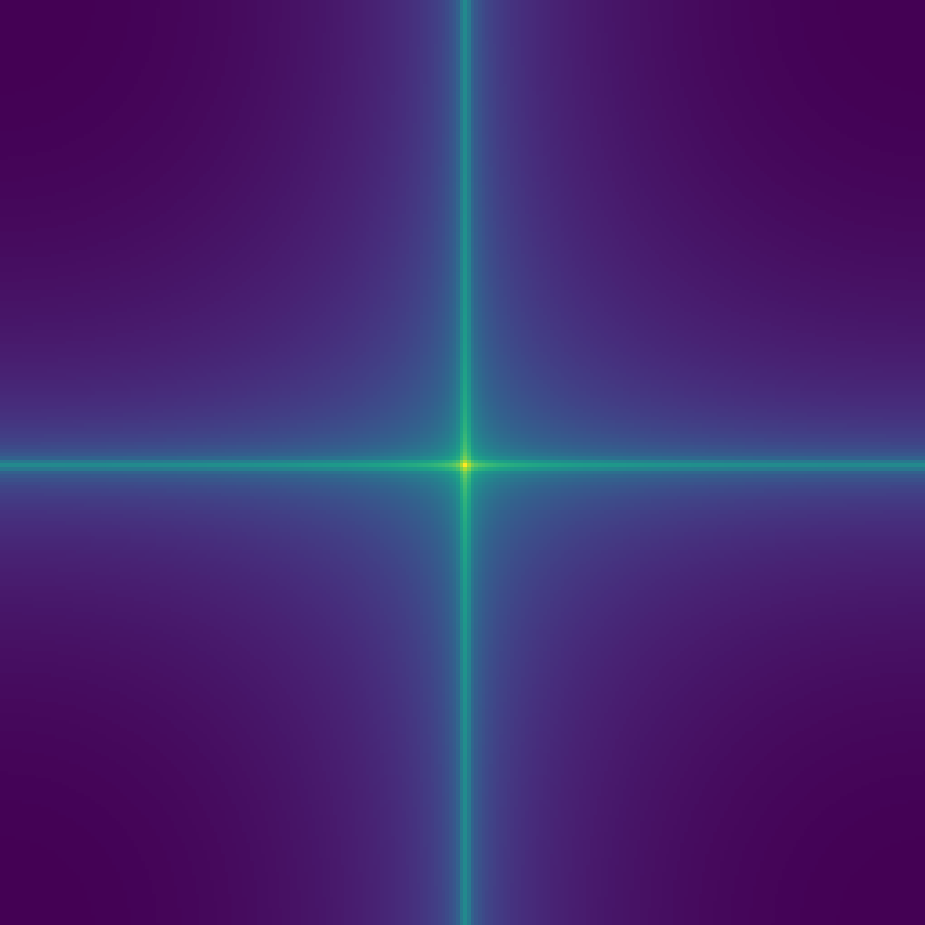} &
        \includegraphics[width=0.49\columnwidth]{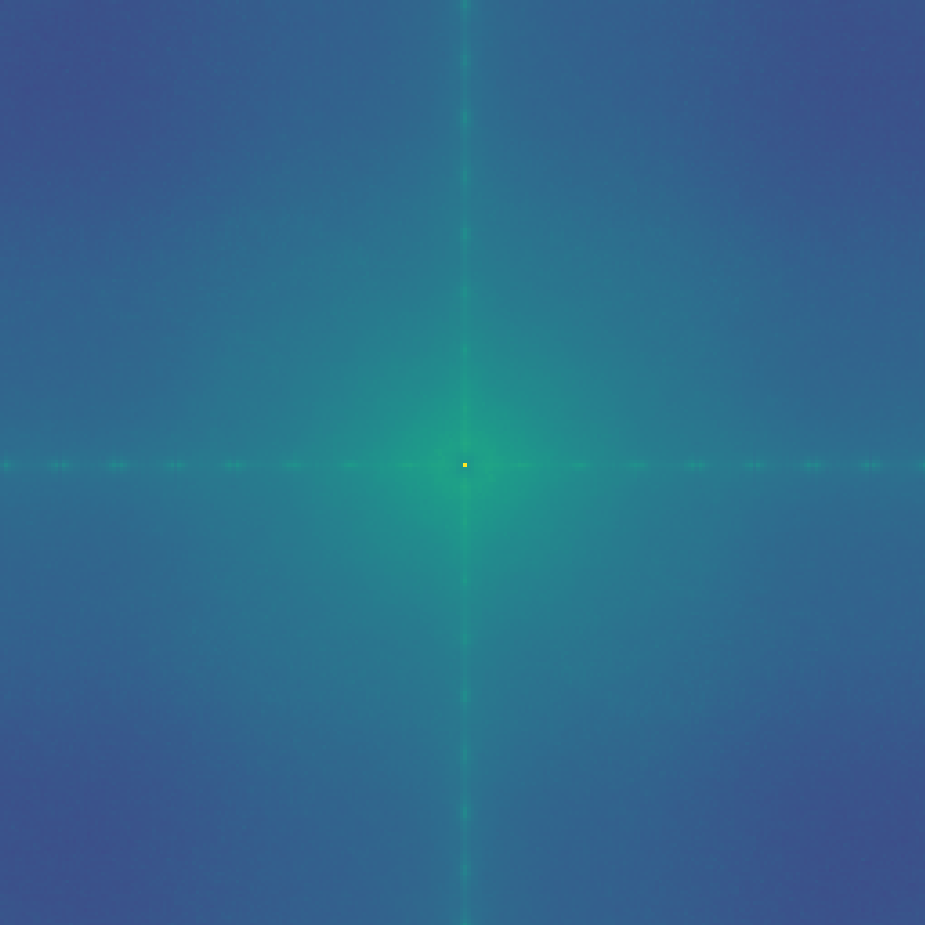}
        \\
        \footnotesize (a) Fourier features &
        \footnotesize (b) Coordinate embeddings
    \end{tabular}
    \caption{The spectrum magnitude for our two kinds of positional encoding (color scale is equal for both plots). The output of coordinate embeddings clearly has more higher frequencies.}
    \label{fig:pos_encoding_magnitude}
\end{figure}

\begin{figure}[ht]
    \centering
    \setlength{\tabcolsep}{2pt}
    \begin{tabular}{cc}
        \includegraphics[width=0.49\columnwidth]{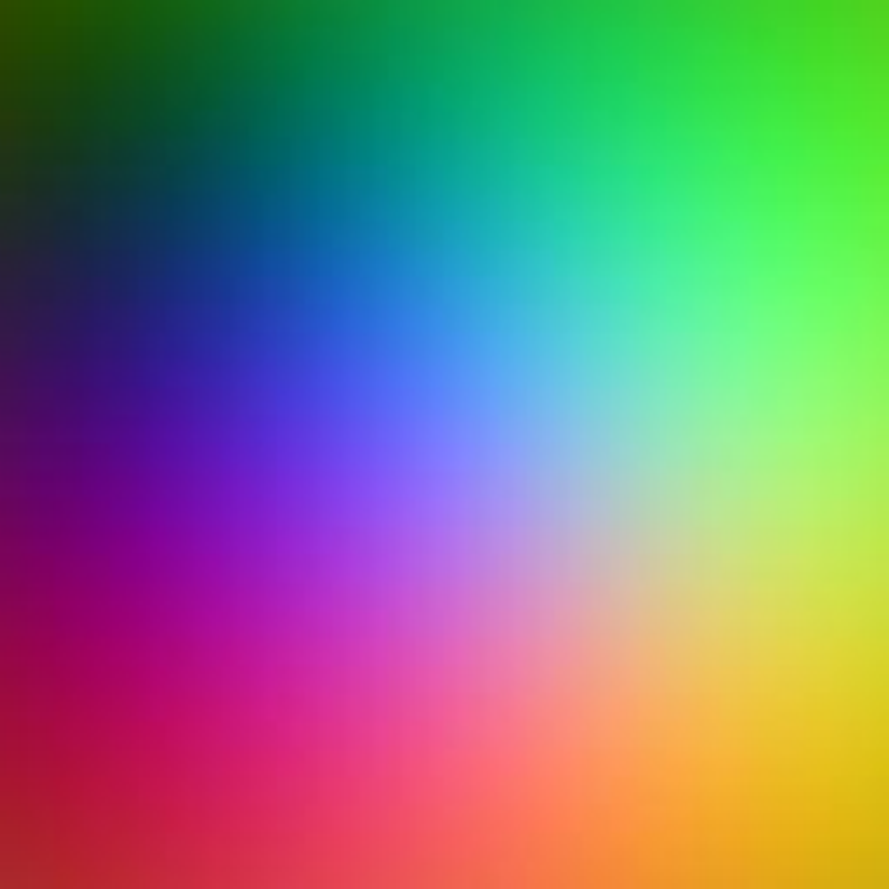} &
        \includegraphics[width=0.49\columnwidth]{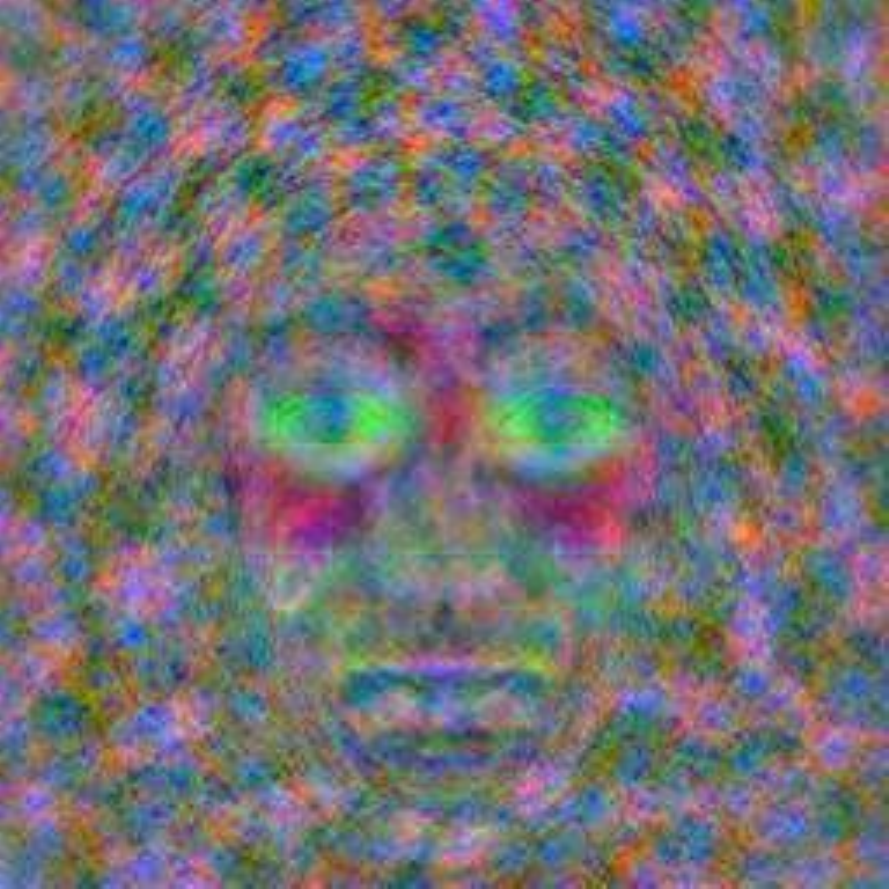}
        \\
        \footnotesize (a) Fourier features &
        \footnotesize (b) Coordinate embeddings
    \end{tabular}
    \caption{PCA plot (3 components) for two kinds of positional encoding of \ganname-base.
        Coordinate embeddings contain not just more fine-grained details, but also key points of the averaged face.
    }
    \label{fig:pca_encoding}
\end{figure}

\begin{figure}[ht]
    \centering
    \setlength{\tabcolsep}{2pt}
    \begin{tabular}{ccc}
        \includegraphics[width=0.31\columnwidth]{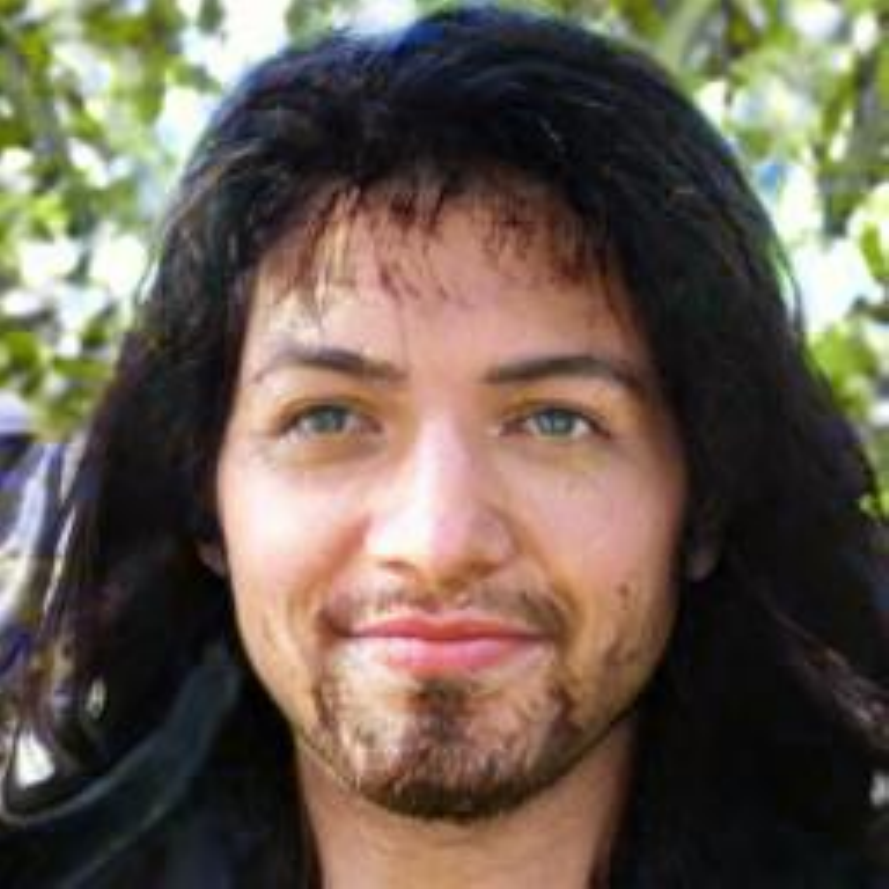} &
        \includegraphics[width=0.31\columnwidth]{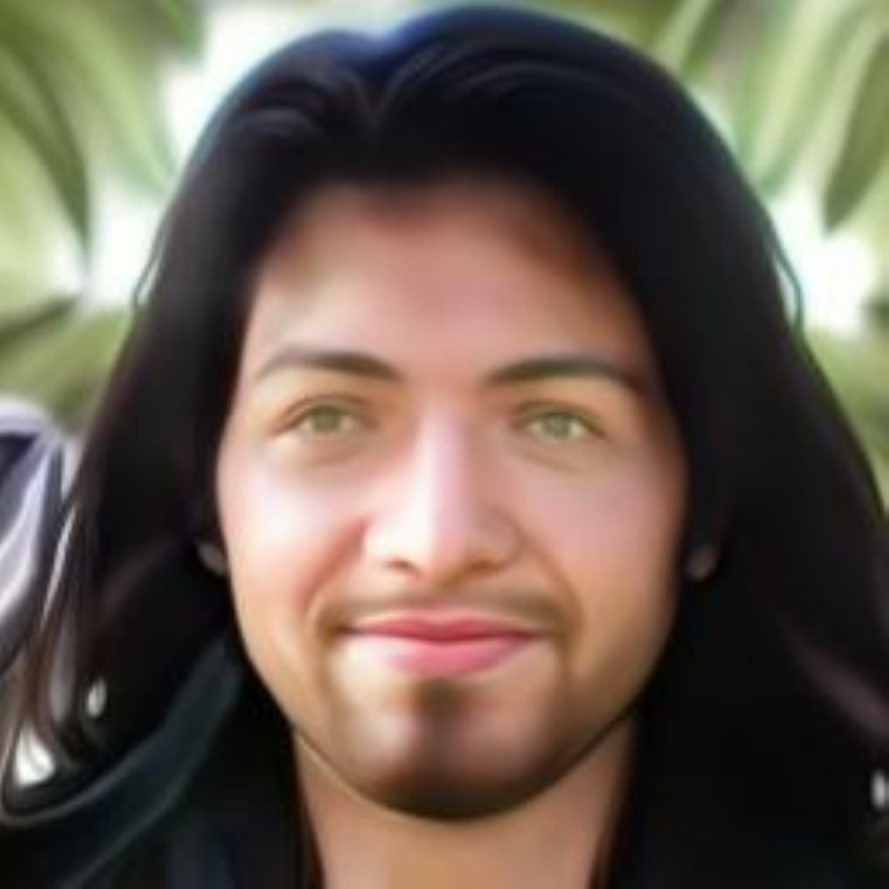} &
        \includegraphics[width=0.31\columnwidth]{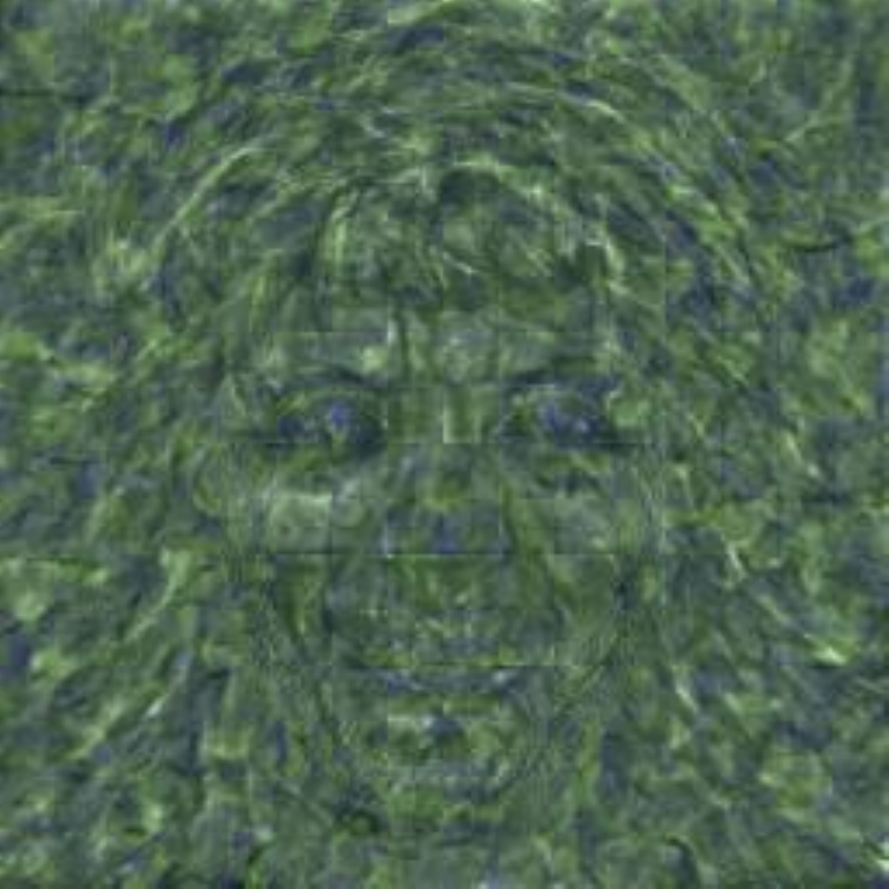}
    \end{tabular}
    \caption{Influence of different types of positional encoding on the resulting image.
        Left: original image.
        Center: coordinate embeddings zeroed out (the image contains no fine-grained details).
        Right: Fourier features zeroed out (only high-frequency details are present). 
    }
    \label{fig:pos_encoding_zeroed}
\end{figure}

To analyze the difference between Fourier features $e_{fo}$ and coordinate embeddings $e_{co}$, we plotted the spectrum of these codes for the generator \ganname-base, trained on FFHQ.
As shown in Fig.~\ref{fig:pos_encoding_magnitude}, Fourier encoding generally carries low-frequency components, whereas coordinate embeddings resemble more high-frequency details.
The Principal Component Analysis (PCA) of the two encodings supports the same conclusion (Fig.~\ref{fig:pca_encoding})
The possible explanation is simple: coordinate embeddings are trained independently for each pixel, while $e_{fo} \left(x, y\right)$ is a learned function of the coordinates.
However, the next layers of the network could transform the positional codes and, for example,  finally produce more fine-grained details, relying on Fourier features.
To demonstrate that this is not the case, we conducted the following experiment. We have zeroed out either the output of Fourier features or coordinate embeddings and showed the obtained images in Fig.~\ref{fig:pos_encoding_zeroed}. One can notice that the information about the facial hair's details as well as the forelock is located in the coordinate embeddings.
This proves that it is the coordinate embeddings that are the key to high-frequency details of the resulting image.

\subsection{Spectral analysis of generated images}
\label{sec:spectral_analysis}

\begin{figure}
\centering
  \begin{subfigure}{\columnwidth}
    \centering
        \includegraphics[width=\linewidth]{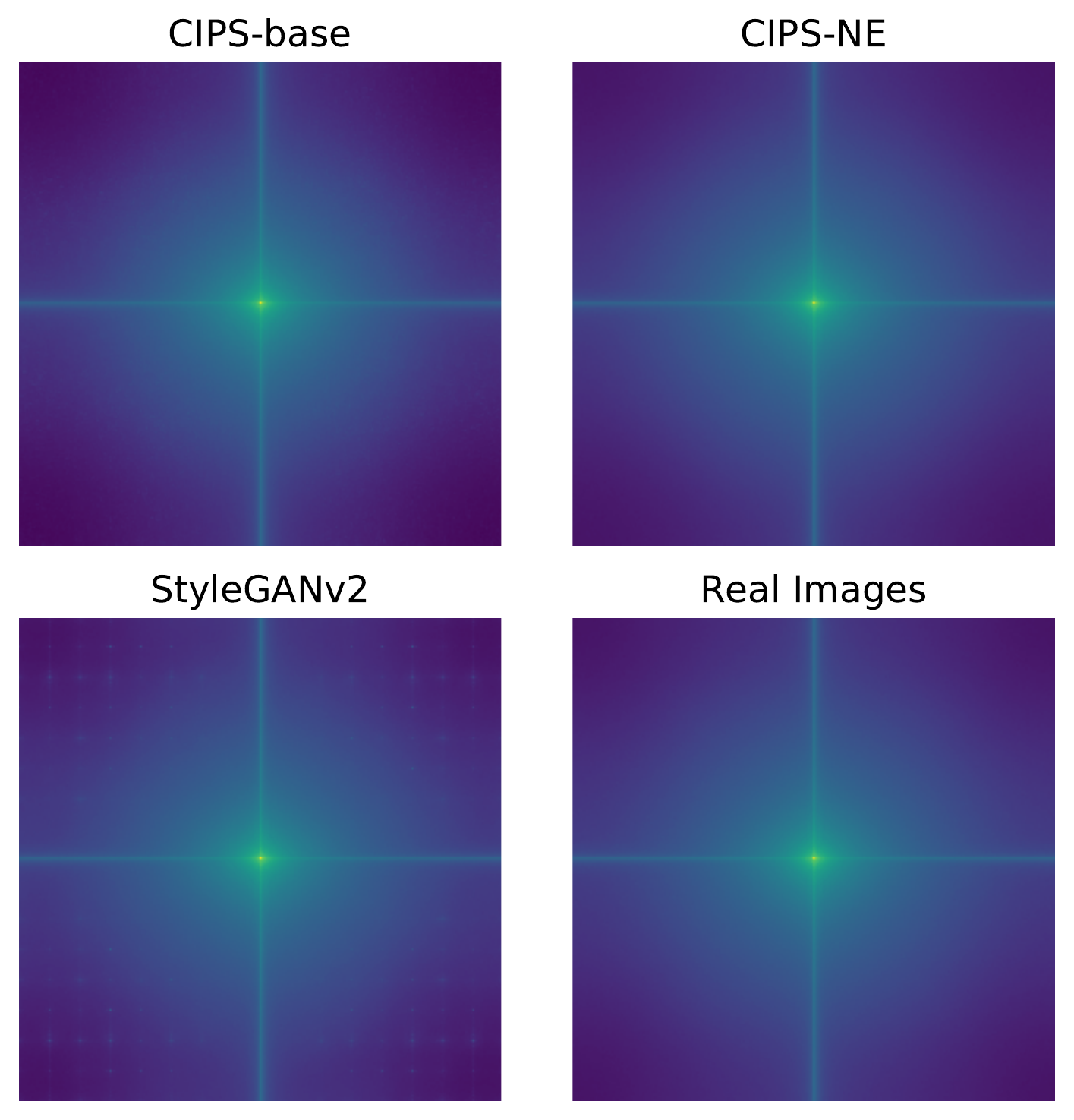}
    \caption{Magnitude spectrum. 
        Our models produce less artifacts in high frequency components (note the grid-like pattern in StyleGANv2). 
        Two \ganname models are difficult to distinguish between (better zoom in).
    }
    \label{fig:2d_spectrum}
\end{subfigure}

\begin{subfigure}{\columnwidth}
 \centering
    \includegraphics[width=\linewidth]{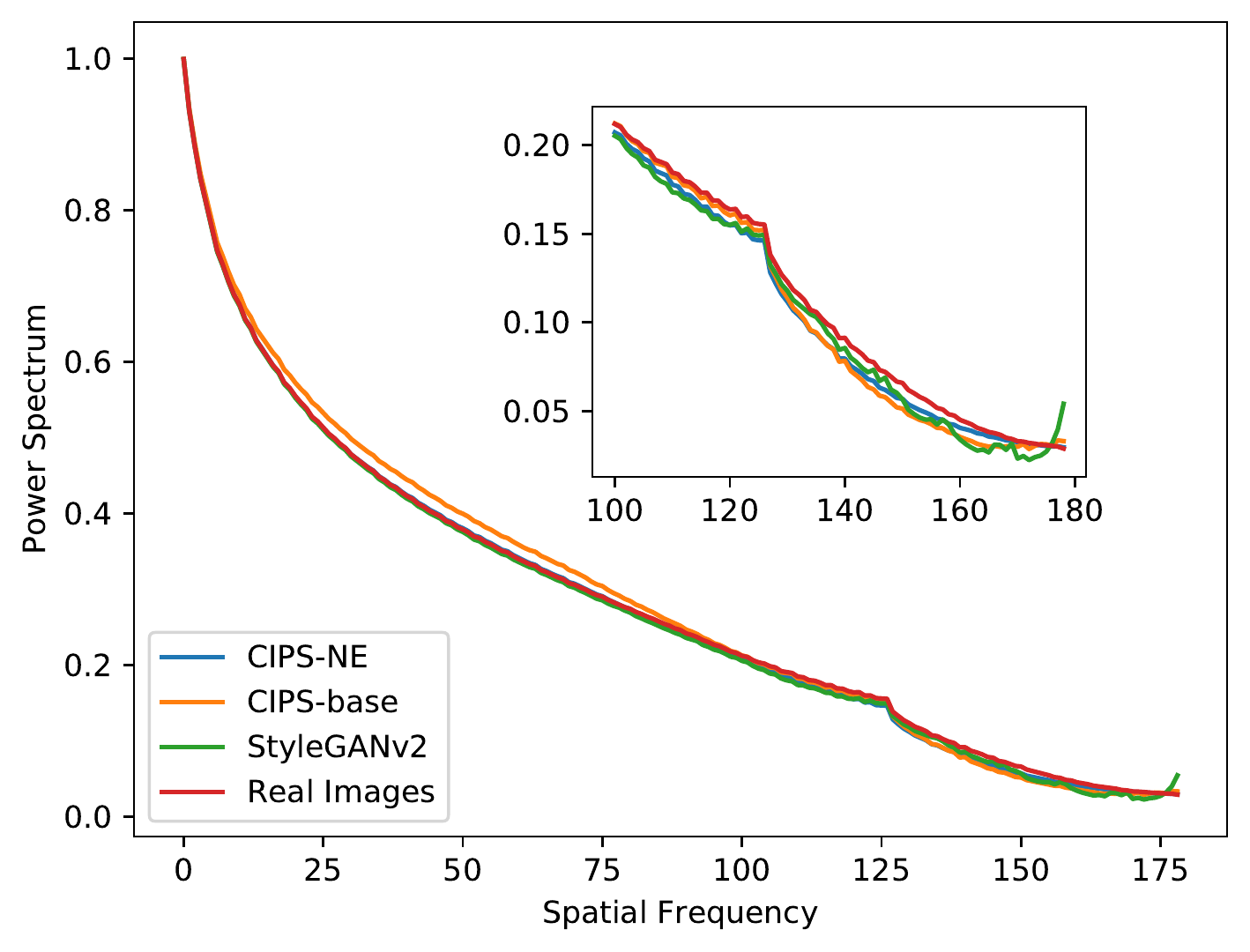}
    \caption{Azimuthal integration over Fourier power spectrum.
        The curve of StyleGANv2 has heavy distortions in most high frequency components.
        Surprisingly, \ganname-NE  demonstrates a more realistic and smooth tail than \ganname-base, while being worse in terms of FID.
    }
    \label{fig:azimuth_psd}
\end{subfigure}
\caption{Spectral analysis for models trained on FFHQ at resolution of $256^2$.
    All results are averaged across 5000 samples. 
    We demonstrate that \ganname-NE is most similar to real images. 
}

\label{fig:spectra}
\end{figure}

Recently, \cite{power_spectral_density} observed that the common convolutional upsampling operations can lead to the inability to learn the spectral distribution of real images, in spite of any generator architecture.
In contrast, \ganname operates explicitly with the coordinate grid and has no upscaling modules, which should to improved reproduction of the spectrum.
Indeed, we compare the spectrum of our models (CIPS-``base'' without residual and skip connections; CIPS-NE) to StyleGANv2 and demonstrate that CIPS generators design has the advantage in the spectral domain. 

The analysis of magnitude spectra for produced images is given in Fig.~\ref{fig:2d_spectrum}.
The spectrum of StyleGANv2 has artifacts in high-frequency regions, not present in both CIPS generators under consideration.
Following prior works~\cite{power_spectral_density}, we also use the azimuthal integration (AI) over the Fourier power spectrum (Fig.~\ref{fig:azimuth_psd}). 
It is worth noting that AI statistics of \ganname-NE are very close to the ones of real images.
However, adding the coordinate embeddings degrades a realistic spectrum while improving the quality in terms of FID~(Tab.~\ref{tab:modifications}).

We note that the introduction  of skip connections in fact makes the spectra less similar to those of natural images.

\subsection{Interpolation}

\begin{figure}[ht!]
    \centering
    \includegraphics[width=\linewidth]{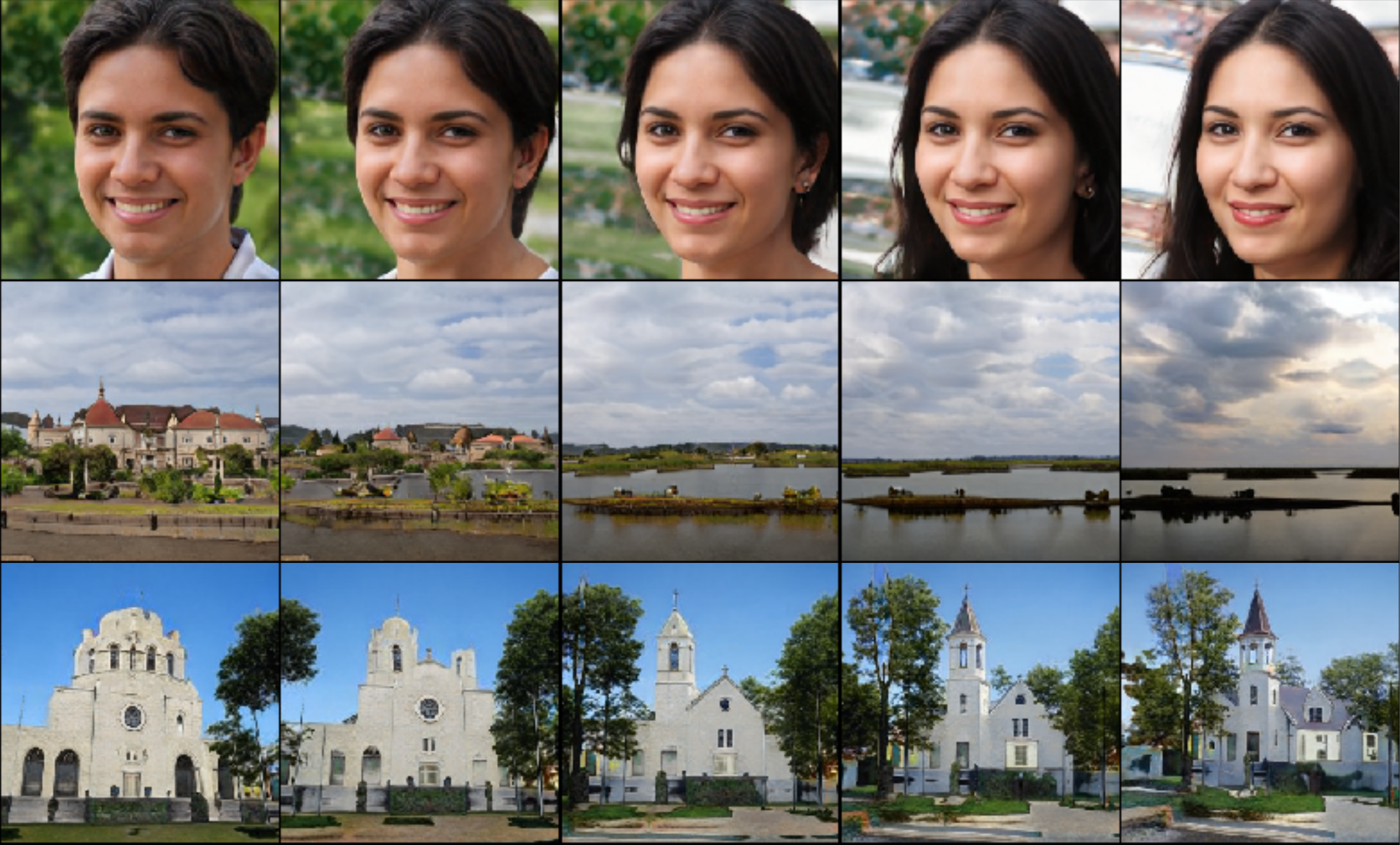}
    \caption{Latent linear morphing between two sampled images -- the left-most and right-most ones.}
    \label{fig:linear_interpolate}
\end{figure}

We conclude the experimental part with the demonstration of the flexibility of \ganname. As well as many other generators, CIPS generators have the ability to interpolate between latent vectors with meaningful morphing (Fig.~\ref{fig:linear_interpolate}). 
As expected, the change between the extreme images occurs smoothly and allows for the use of this property, in a similar vein as in the original works (e.g.~\cite{stylegan2}).

\begin{figure}[t]
    \centering
    \includegraphics[width=\columnwidth]{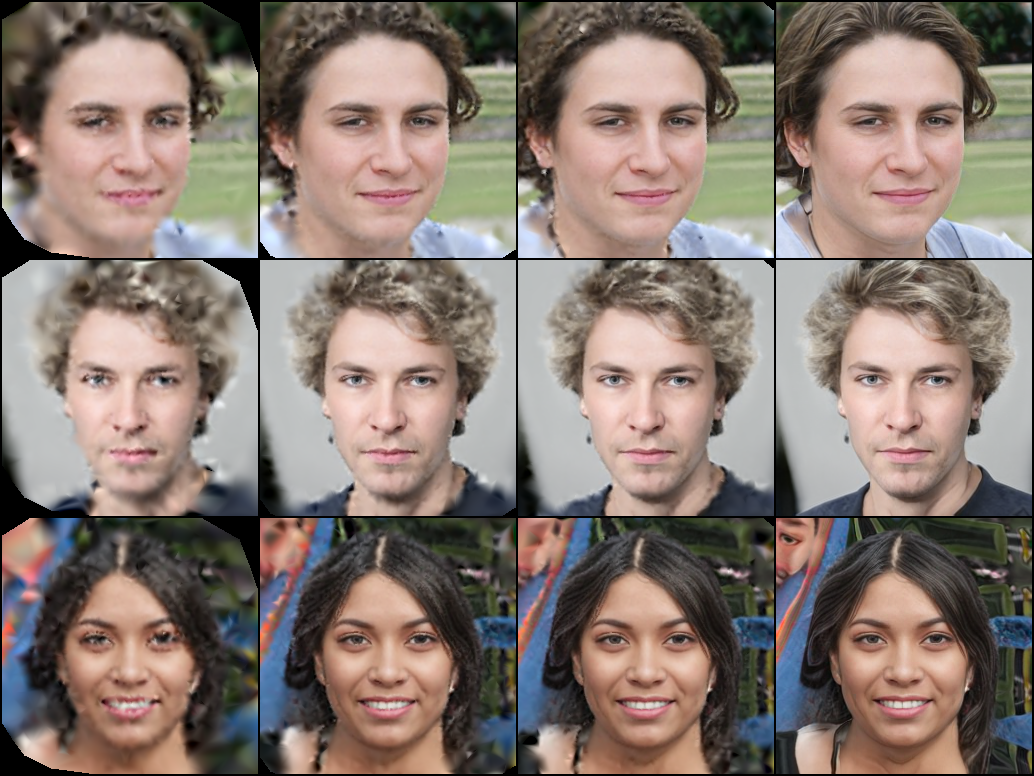}
    \caption{
        Images generated using foveated synthesis. In each case, the CIPS generator was sampled on a 2D Gaussian distribution concentrated in the center of an image (standard deviation = $0.4*$image size). 
        Left to right: sampled pattern covers 5\% of all pixels, 25\%, 50\%, 100\% (full coordinate grid).
        Missing color values have been filled via bicubic interpolation.
    }
    \label{fig:foveated_grid}
\end{figure}

\begin{figure}[t]
    \centering
    \includegraphics[width=\columnwidth]{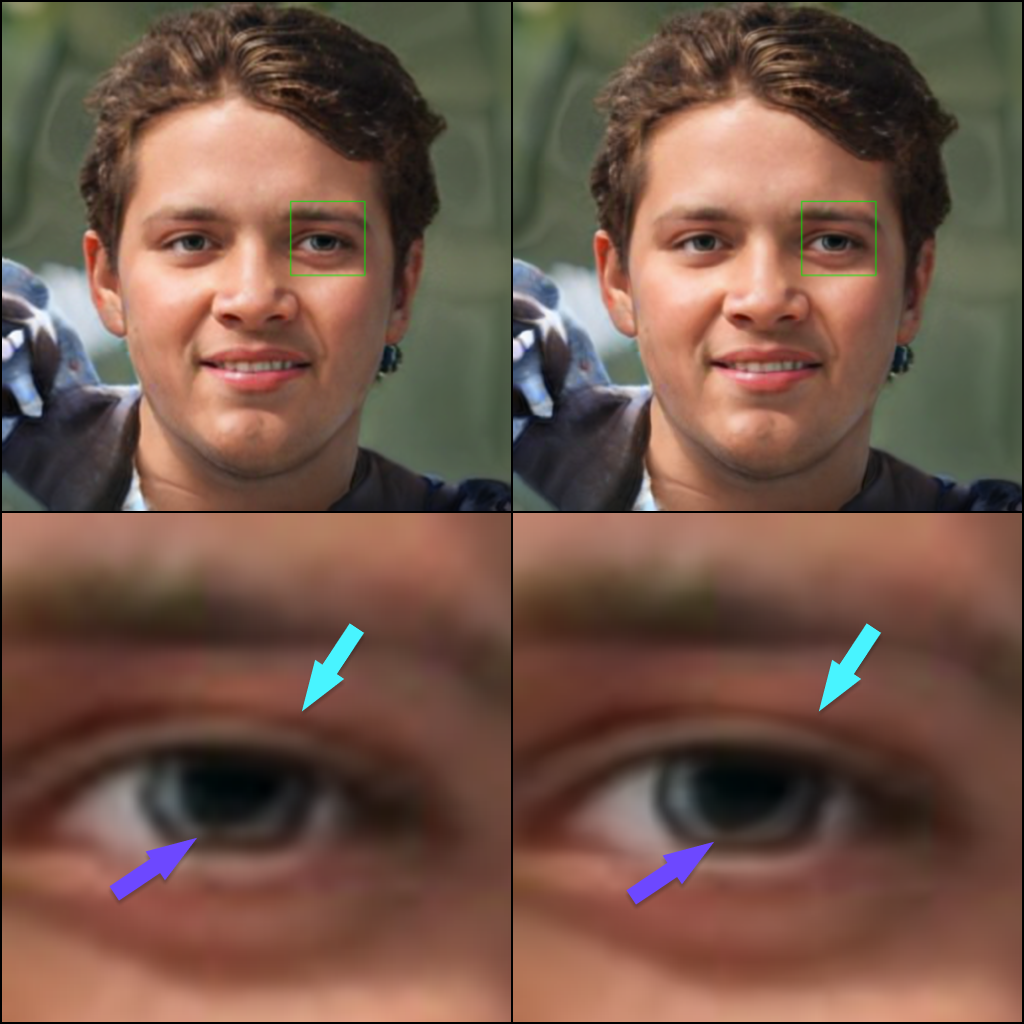}
    \caption{
        Left: the generated image of resolution $256\times{}256$, upscaled with Lanczos upsampling scheme \cite{lanczos1950iteration} to $1024\times{}1024$.
        Right: the image, synthesized by \ganname, trained at resolution of $256\times{}256$ on the coordinate grid of resolution $1024\times{}1024$.
        Note the sharpness/plausibility of the eyelid and the more proper shape of the pupil.
    }
    \label{fig:oversampled_image}
\end{figure}

\subsection{Foveated rendering and interpolation}

One of the inspiring applications of our per-pixel generator is the foveated synthesis. The foveated synthesis ability can be beneficial for computer graphics and other applications, as well as mimics human visual system. In foveated synthesis, an irregular grid of coordinates is sampled first, more dense in the area, where the gaze is assumed to be directed to, and more sparse outside of that region. After that, \ganname is evaluated on this grid (its size is less than the full resolution), and color for missing pixels of the image is filled using interpolation. The demonstration of this method is provided in Fig.~\ref{fig:foveated_grid}.

Alongside the foveated rendering, we are also able to interpolate the image beyond the training resolution by simply sampling denser grids. Here we use a model, trained on images of $256\times{}256$ resolution to process a grid of $1024\times{}1024$ pixels and compare it with upsampling the results of upsampling the image synthesized at the $256\times{}256$ resolution with the Lanczos filter \cite{lanczos1950iteration}. As Fig.~\ref{fig:oversampled_image} suggests, more plausible details are obtained with denser synthesis than with Lanczos filter.

\subsection{Panorama synthesis}

\begin{figure*}[ht!]
\captionsetup[subfigure]{font=footnotesize}

\begin{minipage}{1.\columnwidth}
\centering
\subfloat[]{
    \label{fig:panorama:a}
    \includegraphics[scale=.24]{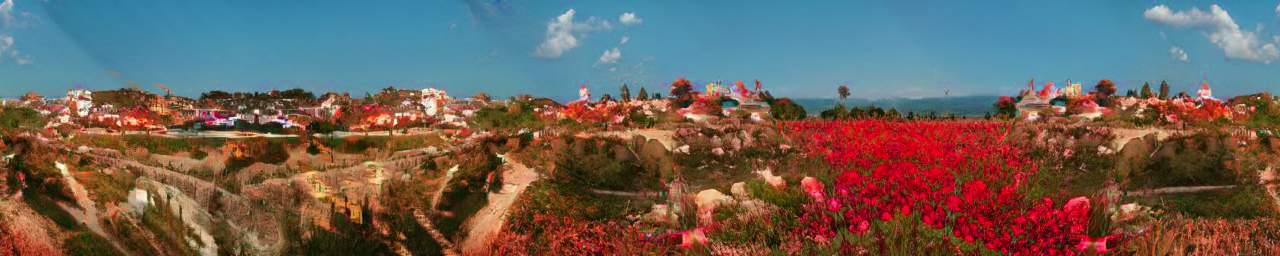}
    }
\end{minipage}%
\hspace{0.9cm}%
\begin{minipage}{1.\columnwidth}
\centering
\subfloat[]{\label{fig:panorama:b}\includegraphics[scale=.24]{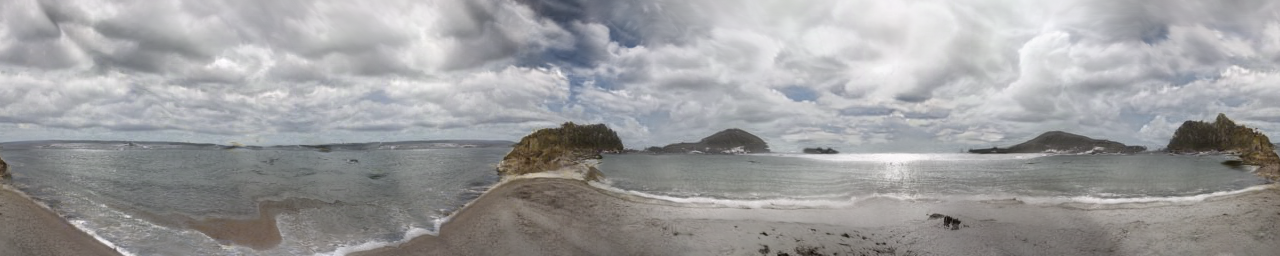}}
\end{minipage}

\par\medskip
\subfloat[]{\label{fig:panorama:result}\includegraphics[scale=.515]{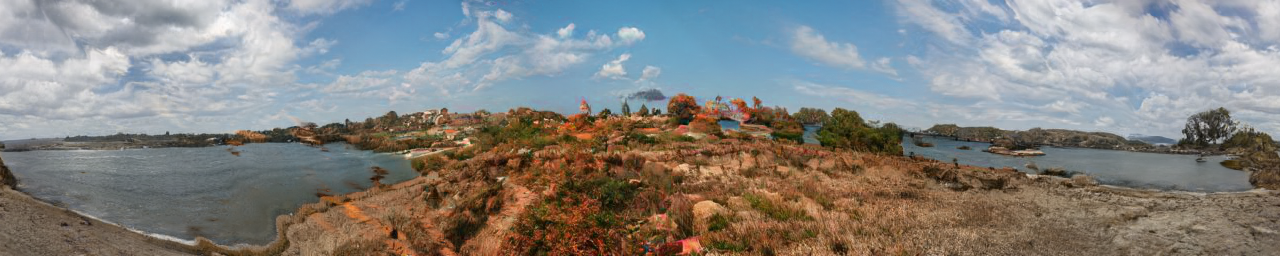}}
\caption{Panorama blending. We linearly blend two upper images from CISP generator trained on the Landscapes dataset with a cylindrical coordinate system. The resulting image contains elements from both original panoramas: land and water integrated naturally.}
\label{fig:panorama}
\end{figure*}

As \ganname is built upon a coordinate grid, it can relatively easily use non-Cartesian grids. To show this, we thus adopt a cylindrical system to produce landscape panoramas.
The training setup is as follows. We uniformly sample a crop $256 \times 256$ from the cylindrical coordinate grid and train the generator to produce images using these coordinate crops as inputs. A similar idea was also explored in \cite{cocogan}. We note, however, that during training we do not use any real panoramas in contrast to other coordinate-based COCO-GAN model~\cite{cocogan}.
Fig.~\ref{fig:panorama:a} and \ref{fig:panorama:b} provide examples of panorama samples obtained with the resulting model.

As each pixel is generated from its coordinates and style vector only, our architecture admits pixel-wise style interpolation (Fig.~\ref{fig:panorama:result}). In these examples, the style vector blends between the central part (the style of Fig.~\ref{fig:panorama:a}) and the outer part (the style of~\ref{fig:panorama:b}).

\subsection{Typical artifacts}

\begin{figure}[ht!]
\centering
\includegraphics[width=\linewidth]{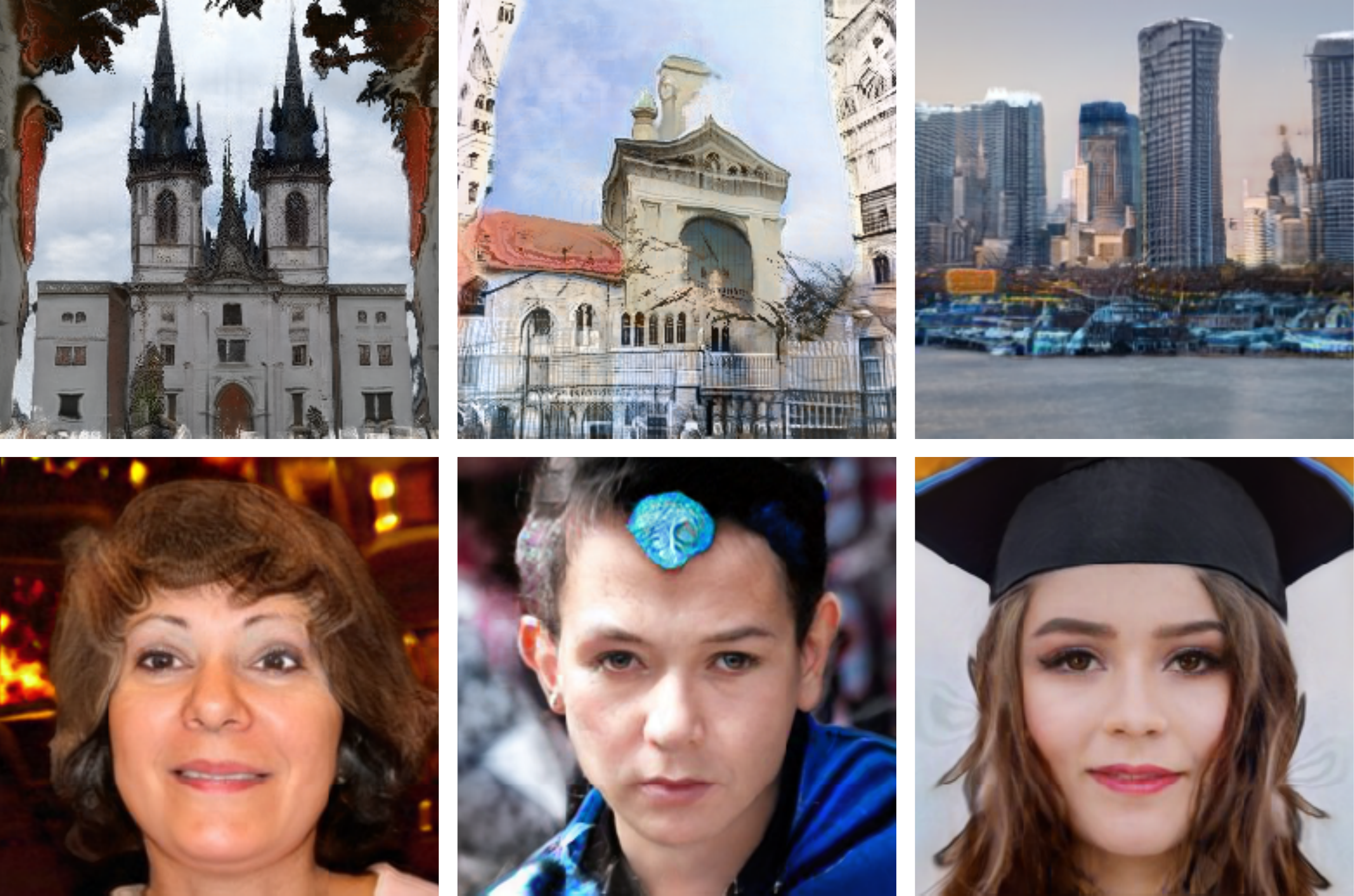}
\caption{Examples of the most  common kinds of artifacts on  different datasets. They are best described as wavy textures on hair, background, and glowing blobs. see text for the discussion. 
}
\label{fig:failure_cassess}
\end{figure}

Finally, we show the typical artifacts that keep recurring in the results of \ganname generators (Fig. \ref{fig:failure_cassess}). We attribute the wavy texture (in hair) and repeated lines pattern (in buildings) to the periodic nature of sine activation function within the Fourier features. Also we note that sometimes \ganname produces a realistic image with a small part of the image being inconsistent with the rest and out of the domain. Our belief is that this behaviour is caused by the LeakyReLU activation function that divides the coordinate grid into parts. For each part, \ganname effectively applies its own inverse discrete  Fourier transform. As \ganname generators do not use any upsampling or other pixel coordination, it is harder for the generator to safeguard against such behaviour.   

\label{sec:failures}

\section{Conclusion}
\label{sec:conclusion}
We presented a new generator model called \ganname, a high-quality architecture with conditionally independent pixel synthesis, such that the color value is computed  using only random noise and coordinate position. 

Our key insight is that the proposed architecture without spatial convolutions, attention or upsampling operations has the ability to compete in the model market and obtain decent quality in terms of FID and precision \& recall; such results have not been presented earlier for perceptron-based models. 
Furthermore, in the spectral domain  outputs of  \ganname are harder to discriminate from real images.
Interestingly, \ganname-NE modification is weaker in terms of plausibility, yet has a more realistic spectrum.

Direct usage of a coordinate grid allows us to work with more complex structures, such as cylindrical panoramas, just by replacing the underlying coordinate system. 

In summary, our generator demonstrates quality on par with state-of-the-art model StyleGANv2; moreover, it has applications in various diverse scenarios.
We have shown that the considered model could be successfully applied to foveated rendering and super-resolution problems in their generative interpretations. 
Future development of our approach assumes researching these problems in their image-to-image formulations.

{\small
\bibliographystyle{ieee}
\bibliography{refs}

\begin{thebibliography}{10}\itemsep=-1pt

\bibitem{biggan}
A.~Brock, J.~Donahue, and K.~Simonyan.
\newblock Large scale {GAN} training for high fidelity natural image synthesis.
\newblock In {\em International Conference on Learning Representations}, 2019.

\bibitem{cohen2018spherical}
T.~S. Cohen, M.~Geiger, J.~K{\"o}hler, and M.~Welling.
\newblock Spherical cnns.
\newblock In {\em International Conference on Learning Representations}, 2018.

\bibitem{Dinh16}
L.~Dinh, J.~Sohl-Dickstein, and S.~Bengio.
\newblock Density estimation using real nvp.
\newblock {\em arXiv preprint arXiv:1605.08803}, 2016.

\bibitem{power_spectral_density}
R.~{Durall}, M.~{Keuper}, and J.~{Keuper}.
\newblock {W}atch {Y}our {U}p-{C}onvolution: {CNN} {B}ased {G}enerative {D}eep
  {N}eural {N}etworks {A}re {F}ailing to {R}eproduce {S}pectral
  {D}istributions.
\newblock In {\em Proc. {CVPR}}, pages 7887--7896, 2020.

\bibitem{otoro1}
D.~Ha.
\newblock Generating large images from latent vectors.
\newblock {\em blog.otoro.net}, 2016.

\bibitem{otoro2}
D.~Ha.
\newblock Generating large images from latent vectors - part two.
\newblock {\em blog.otoro.net}, 2016.

\bibitem{fid}
M.~Heusel, H.~Ramsauer, T.~Unterthiner, B.~Nessler, and S.~Hochreiter.
\newblock Gans trained by a two time-scale update rule converge to a local nash
  equilibrium.
\newblock In {\em Proc. {NIPS}}, NIPS'17, page 6629–6640, Red Hook, NY, USA,
  2017. Curran Associates Inc.

\bibitem{karpathy_regression}
A.~Karpathy.
\newblock Convnetjs demo: Image "painting".
\newblock
  \url{https://cs.stanford.edu/people/karpathy/convnetjs/demo/image_regression.html}.
\newblock Accessed: 2020-11-05.

\bibitem{aug_nvidia}
T.~Karras, M.~Aittala, J.~Hellsten, S.~Laine, J.~Lehtinen, and T.~Aila.
\newblock Training {Generative} {Adversarial} {Networks} with {Limited} {Data}.
\newblock In {\em Proc. {NeurIPS}}. Curran Associates, Inc., 2020.

\bibitem{stylegan1}
T.~{Karras}, S.~{Laine}, and T.~{Aila}.
\newblock A style-based generator architecture for generative adversarial
  networks.
\newblock In {\em Proc. {CVPR}}, pages 4396--4405, 2019.

\bibitem{stylegan2}
T.~{Karras}, S.~{Laine}, M.~{Aittala}, J.~{Hellsten}, J.~{Lehtinen}, and
  T.~{Aila}.
\newblock Analyzing and improving the image quality of stylegan.
\newblock In {\em Proc. {CVPR}}, pages 8107--8116, 2020.

\bibitem{adam}
D.~P. Kingma and J.~Ba.
\newblock Adam: {A} method for stochastic optimization.
\newblock In {\em International Conference on Learning Representations,
  {ICLR}}, 2015.

\bibitem{Kingma18}
D.~P. Kingma and P.~Dhariwal.
\newblock Glow: Generative flow with invertible 1x1 convolutions.
\newblock In {\em Proc. {NeurIPS}}, pages 10215--10224, 2018.

\bibitem{kingma2013auto}
D.~P. Kingma and M.~Welling.
\newblock Auto-encoding variational bayes.
\newblock {\em arXiv preprint arXiv:1312.6114}, 2013.

\bibitem{improved_prec_rec}
T.~Kynk\"{a}\"{a}nniemi, T.~Karras, S.~Laine, J.~Lehtinen, and T.~Aila.
\newblock Improved precision and recall metric for assessing generative models.
\newblock In H.~Wallach, H.~Larochelle, A.~Beygelzimer, F.~d\textquotesingle
  Alch\'{e}-Buc, E.~Fox, and R.~Garnett, editors, {\em Proc. {NeurIPS}},
  volume~32, pages 3927--3936. Curran Associates, Inc., 2019.

\bibitem{lanczos1950iteration}
C.~Lanczos.
\newblock {\em An iteration method for the solution of the eigenvalue problem
  of linear differential and integral operators}.
\newblock United States Governm. Press Office Los Angeles, CA, 1950.

\bibitem{cocogan}
C.~H. {Lin}, C.~{Chang}, Y.~{Chen}, D.~{Juan}, W.~{Wei}, and H.~{Chen}.
\newblock Coco-gan: Generation by parts via conditional coordinating.
\newblock In {\em Proc. {ICCV}}, pages 4511--4520, 2019.

\bibitem{coordconv}
R.~Liu, J.~Lehman, P.~Molino, F.~Petroski~Such, E.~Frank, A.~Sergeev, and
  J.~Yosinski.
\newblock An intriguing failing of convolutional neural networks and the
  {CoordConv} solution.
\newblock In S.~Bengio, H.~Wallach, H.~Larochelle, K.~Grauman, N.~Cesa-Bianchi,
  and R.~Garnett, editors, {\em Proc. {NeurIPS}}, pages 9627--9638. Curran
  Associates, Inc., 2018.

\bibitem{r1_penalty}
L.~Mescheder, A.~Geiger, and S.~Nowozin.
\newblock Which {Training} {Methods} for {GANs} do actually {Converge}?
\newblock In J.~Dy and A.~Krause, editors, {\em Proc. {ICML}}, volume~80 of
  {\em Proceedings of {Machine} {Learning} {Research}}, pages 3481--3490,
  Stockholmsmässan, Stockholm Sweden, July 2018. PMLR.

\bibitem{nerf}
B.~Mildenhall, P.~P. Srinivasan, M.~Tancik, J.~T. Barron, R.~Ramamoorthi, and
  R.~Ng.
\newblock {NeRF}: {Representing} {Scenes} as {Neural} {Radiance} {Fields} for
  {View} {Synthesis}.
\newblock In A.~Vedaldi, H.~Bischof, T.~Brox, and J.-M. Frahm, editors, {\em
  Proc. {ECCV}}, pages 405--421, Cham, 2020. Springer International Publishing.

\bibitem{mordvintsev_xy2rgb}
A.~Mordvintsev, N.~Pezzotti, L.~Schubert, and C.~Olah.
\newblock Differentiable image parameterizations.
\newblock {\em Distill}, 2018.
\newblock https://distill.pub/2018/differentiable-parameterizations.

\bibitem{dcgan}
A.~Radford, L.~Metz, and S.~Chintala.
\newblock Unsupervised representation learning with deep convolutional
  generative adversarial networks.
\newblock In {\em International Conference on Learning Representations}, 2016.

\bibitem{precision_recall}
M.~S.~M. Sajjadi, O.~Bachem, M.~Lucic, O.~Bousquet, and S.~Gelly.
\newblock Assessing generative models via precision and recall.
\newblock In S.~Bengio, H.~Wallach, H.~Larochelle, K.~Grauman, N.~Cesa-Bianchi,
  and R.~Garnett, editors, {\em Proc. {NIPS}}, volume~31, pages 5228--5237.
  Curran Associates, Inc., 2018.

\bibitem{facenet}
F.~{Schroff}, D.~{Kalenichenko}, and J.~{Philbin}.
\newblock Facenet: A unified embedding for face recognition and clustering.
\newblock In {\em 2015 IEEE Conference on Computer Vision and Pattern
  Recognition (CVPR)}, pages 815--823, 2015.

\bibitem{graf}
K.~Schwarz, Y.~Liao, M.~Niemeyer, and A.~Geiger.
\newblock {GRAF}: {Generative} {Radiance} {Fields} for {3D}-{Aware} {Image}
  {Synthesis}.
\newblock In {\em Proc. {NeurIPS}}. Curran Associates, Inc., 2020.

\bibitem{siren}
V.~Sitzmann, J.~N.~P. Martel, A.~W. Bergman, D.~B. Lindell, and G.~Wetzstein.
\newblock {I}mplicit {N}eural {R}epresentations with {P}eriodic {A}ctivation
  {F}unctions.
\newblock In {\em Proc. {NeurIPS}}. Curran Associates, Inc., 2020.

\bibitem{srn}
V.~Sitzmann, M.~Zollh{\"o}fer, and G.~Wetzstein.
\newblock Scene representation networks: Continuous 3d-structure-aware neural
  scene representations.
\newblock In {\em Proc. {NeurIPS}}. 2019.

\bibitem{cppn}
K.~O. Stanley.
\newblock Compositional pattern producing networks: A novel abstraction of
  development.
\newblock {\em Genetic Programming and Evolvable Machines}, 8(2):131--162, Jun
  2007.

\bibitem{fourier_features}
M.~Tancik, P.~P. Srinivasan, B.~Mildenhall, S.~Fridovich-Keil, N.~Raghavan,
  U.~Singhal, R.~Ramamoorthi, J.~T. Barron, and R.~Ng.
\newblock Fourier {Features} {Let} {Networks} {Learn} {High} {Frequency}
  {Functions} in {Low} {Dimensional} {Domains}.
\newblock In {\em Proc. {NeurIPS}}. Curran Associates, Inc., 2020.

\bibitem{van2016pixel}
A.~Van Den~Oord, N.~Kalchbrenner, and K.~Kavukcuoglu.
\newblock Pixel recurrent neural networks.
\newblock In {\em Proc. {ICML}}, pages 1747--1756, 2016.

\bibitem{lsun}
F.~Yu, A.~Seff, Y.~Zhang, S.~Song, T.~Funkhouser, and J.~Xiao.
\newblock {LSUN}: {C}onstruction of a {L}arge-scale {I}mage {D}ataset using
  {D}eep {L}earning with {H}umans in the {L}oop.
\newblock 2016.

\bibitem{sagan}
H.~Zhang, I.~J. Goodfellow, D.~N. Metaxas, and A.~Odena.
\newblock Self-attention generative adversarial networks.
\newblock In {\em Proc. {ICML}}, 2019.

\bibitem{mtcnn}
K.~{Zhang}, Z.~{Zhang}, Z.~{Li}, and Y.~{Qiao}.
\newblock Joint face detection and alignment using multitask cascaded
  convolutional networks.
\newblock {\em IEEE Signal Processing Letters}, 23(10):1499--1503, 2016.

\bibitem{aug_mit}
S.~Zhao, Z.~Liu, J.~Lin, J.-Y. Zhu, and S.~Han.
\newblock Differentiable {Augmentation} for {Data}-{Efficient} {GAN}
  {Training}.
\newblock In {\em Proc. {NeurIPS}}. Curran Associates, Inc., 2020.

\end{thebibliography}
}
\cleardoublepage
\begin{appendices}
\section{Architecture details}

 \begin{figure}[h!]
    \centering
    \includegraphics[width=0.85\linewidth]{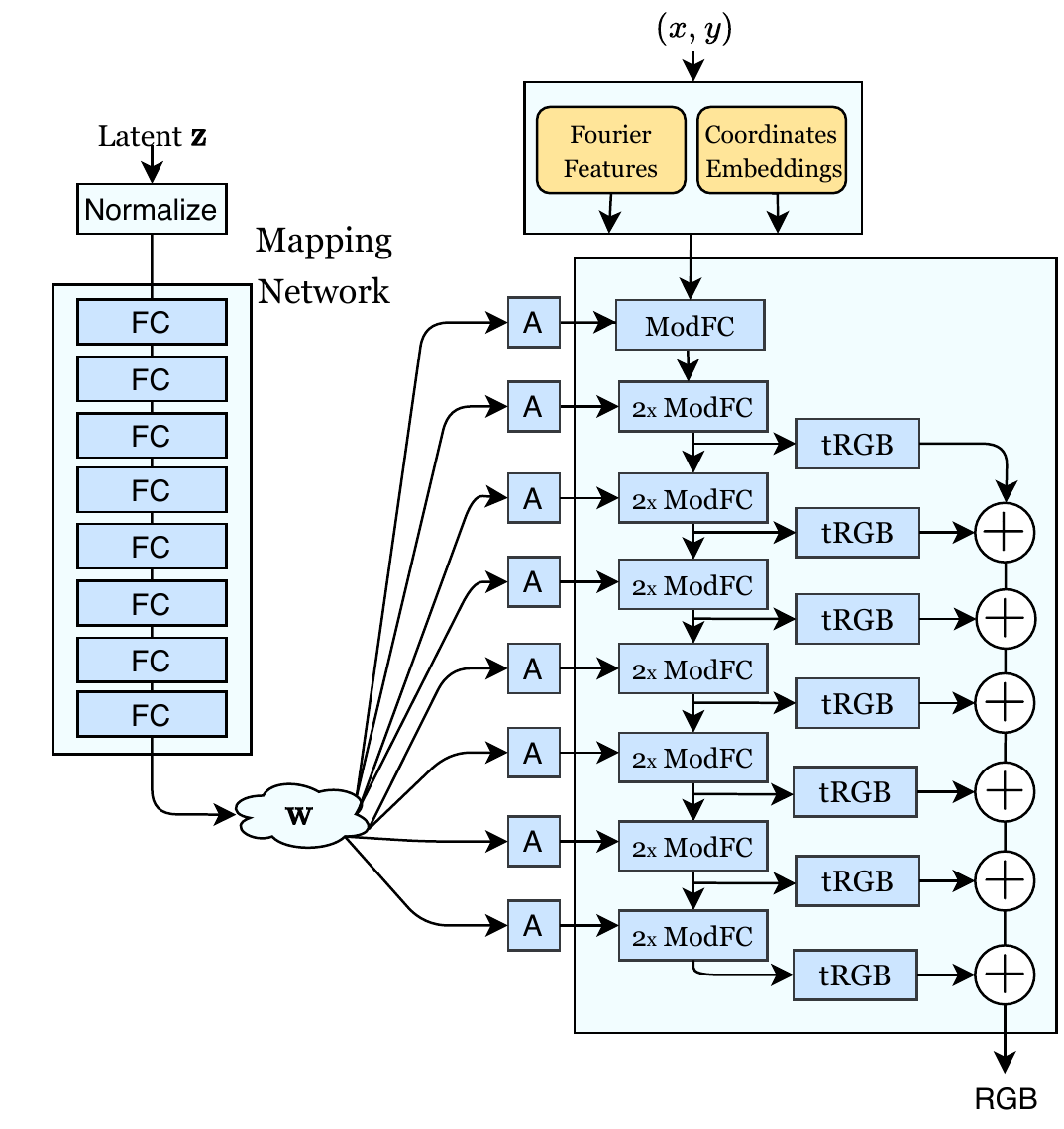}
    \caption{The diagram of the CIPS generator (default version).}
    \label{fig:cips_scheme}
\end{figure}

\begin{table}[h!]
    \centering
    \begin{tabular}{|l|c|} 
        \hline
        Modification & \# parameters (mln) \\
        \hline
        CIPS-``base''  & 43.8 \\
        CIPS-NE & 10.2 \\
        CIPS-Default & 45.9 \\
        \hline
        StyleGANv2 & 30.0 \\
        \hline
    \end{tabular}
    \caption{The number of parameters for different version of the CIPS generator.
        For reference, the number of parameters within the StyleGANv2 generator is also given.
    }
    \label{tab:num_params}
\end{table}

In this section we provide additional information about the default version of our CIPS generator (Fig.~\ref{fig:cips_scheme}). In total, its backbone contains 15 fully connected layers. The first layer projects concatenated coordinate embeddings and Fourier features into the joint space with the dimension of 512. 
Next, the following layer pattern is repeated seven times. The representation is put through two modulated fully-connected layers and a projection to RGB color space is computed. The projections coming from the seven iterations are summed together to create the final image.
The number of parameters for the different modifications of the CIPS generator discussed in the paper are given in Tab.~\ref{tab:num_params}.

\section{Coordinate embeddings}
\label{sec:coordinate_embeddings}

\begin{figure}[t!]
    \centering
    \setlength{\tabcolsep}{2pt}
    \begin{tabular}{ccc}
        \includegraphics[width=0.31\columnwidth]{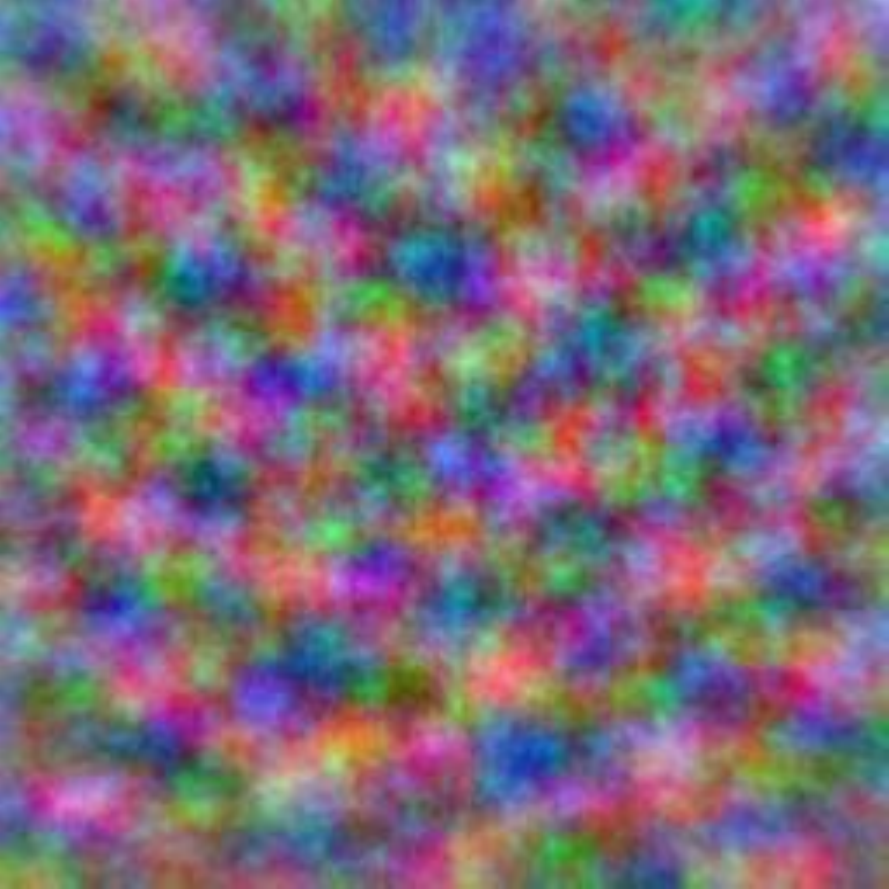} &
        \includegraphics[width=0.31\columnwidth]{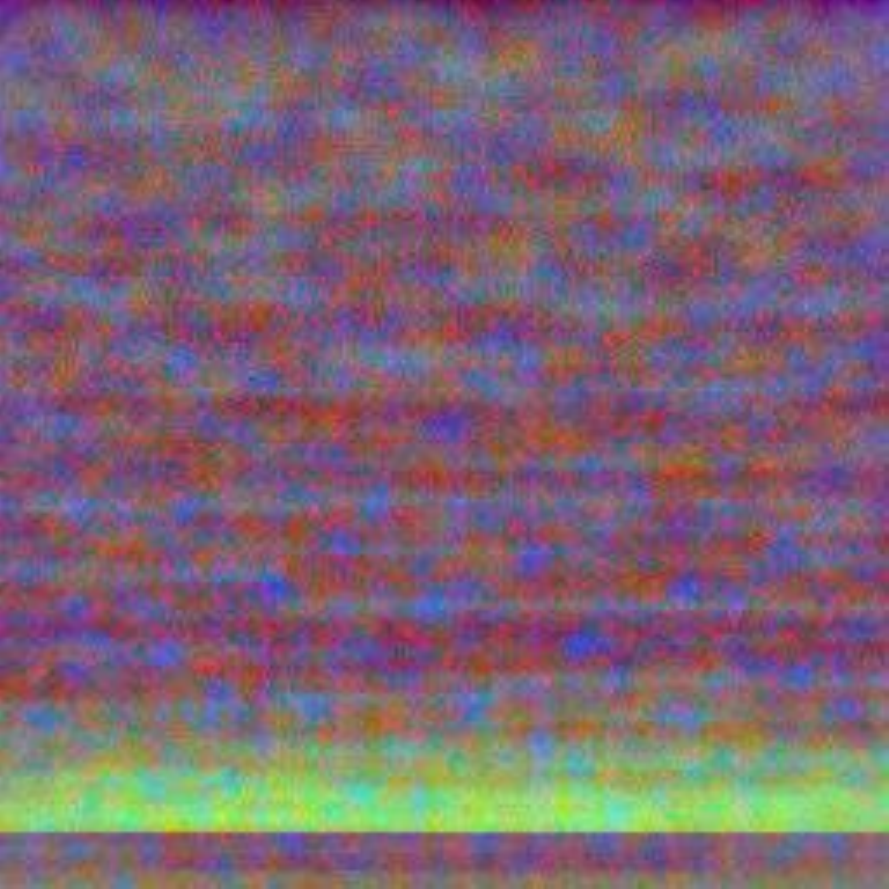} &
        \includegraphics[width=0.31\columnwidth]{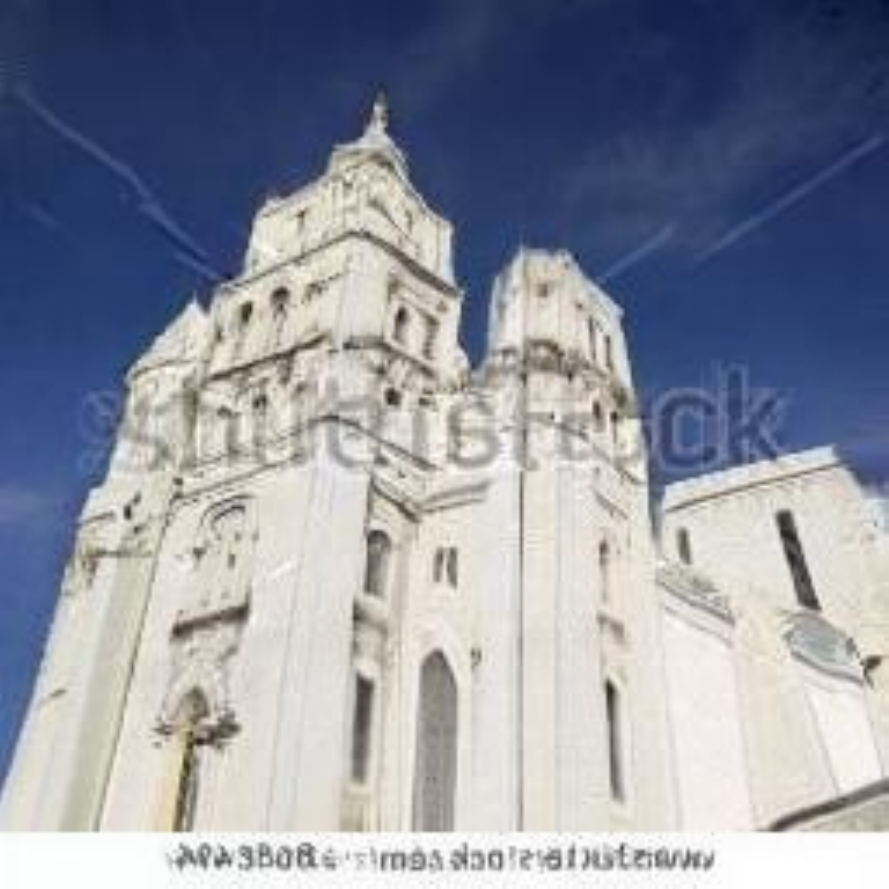}
    \end{tabular}
    \caption{Visualisation of three main principal components of coordinate embeddings for CIPS models, trained on Landscapes (left) and LSUN-Churches (center). As these datasets are not as aligned as the face dataset, there is less recognizable structure in the learned coordinate embeddings. The bottom horizontal structure in the LSUN-Churches case is likely due to frequent watermark pattern in the dataset (a sample from the model with such watermark is shown on the right).
    }
    \label{fig:pca_churches}
    \vspace{-1ex}
\end{figure}

We also run the Principle Components Analysis (PCA) for coordinate embeddings of models trained on Landscapes and LSUN-Churches images (similar pattern for the FFHQ dataset is shown in the main paper).
Fig.~\ref{fig:pca_churches} provides the visualisation for the three main components. Note, that as these datasets are as aligned as FFHQ, there is considerably less spatial structural information in the learned embeddings.

\section{Patch-based generation}
\label{sec:patch_based}

\begin{table}[t]
    \centering
    \begin{tabular}{|l|c|c|} 
        \hline
        Dataset & Patch size & FID \\
        \hline
        \multirow{3}{7em}{FFHQ} & 256 & 4.38 \\
        & 128 & 9.08 \\
        & 64 & 11.79 \\
        \hline
        \multirow{3}{7em}{LSUN-Churches} & 256 & 2.92 \\
        & 128 & 7.08 \\
        & 64 & 11.53 \\
        \hline
    \end{tabular}
    \caption{Frechet Inception Distance (FID) values for CIPS models trained on patches of varying receptive field and fixed resolution ($64\times{}64$ and $128\times{}128$). The results for patch-based training are worse than the default training procedure, in which the discriminator observes the full $256\times 256$ image.
    }
    \label{tab:patches}
\end{table}

To show one benefit of coordinate-based approach, we demonstrate the results of \textit{memory-constrained} training, where the discriminator observes patches at lower resolution than the full image (inspired by the GRAF system~\cite{graf}). Since pixel generation is conditionally-independent, at each iteration only low-resolution patches need to be generated. Thus, only the following $K \times K$ patch is synthesized and submitted to the discriminator:
\small
\begin{align*}
    P_{K, \sigma} \left( u, v \right) =  \lbrace G\left(u+i\sigma, v+j\sigma; \mathbf{z}\right) \mid \left(i, j\right) \in \texttt{mgrid}\left(K, K \right)\rbrace,
    \label{eq:patch_gen}
\end{align*}
\normalsize
where $0 \le u < W-\left(K-1\right)\sigma$ and $0 \le v < H-\left(K-1\right)\sigma$ are the coordinates of the corner pixel of the patch.
For $\sigma = 1$ this produces dense patch, while for $\sigma > 1$ dilated patch with increased receptive field is obtained. Applying this patch sampling to real images before putting them into the discriminator may be thought of as an example of a differentiable augmentation, the usefulness of which was recently proved by~\cite{aug_nvidia,aug_mit}.

\begin{figure}[t!]
    \centering
    
    \begin{subfigure}[b]{0.45\textwidth}
    \centering
        \includegraphics[width=\linewidth]{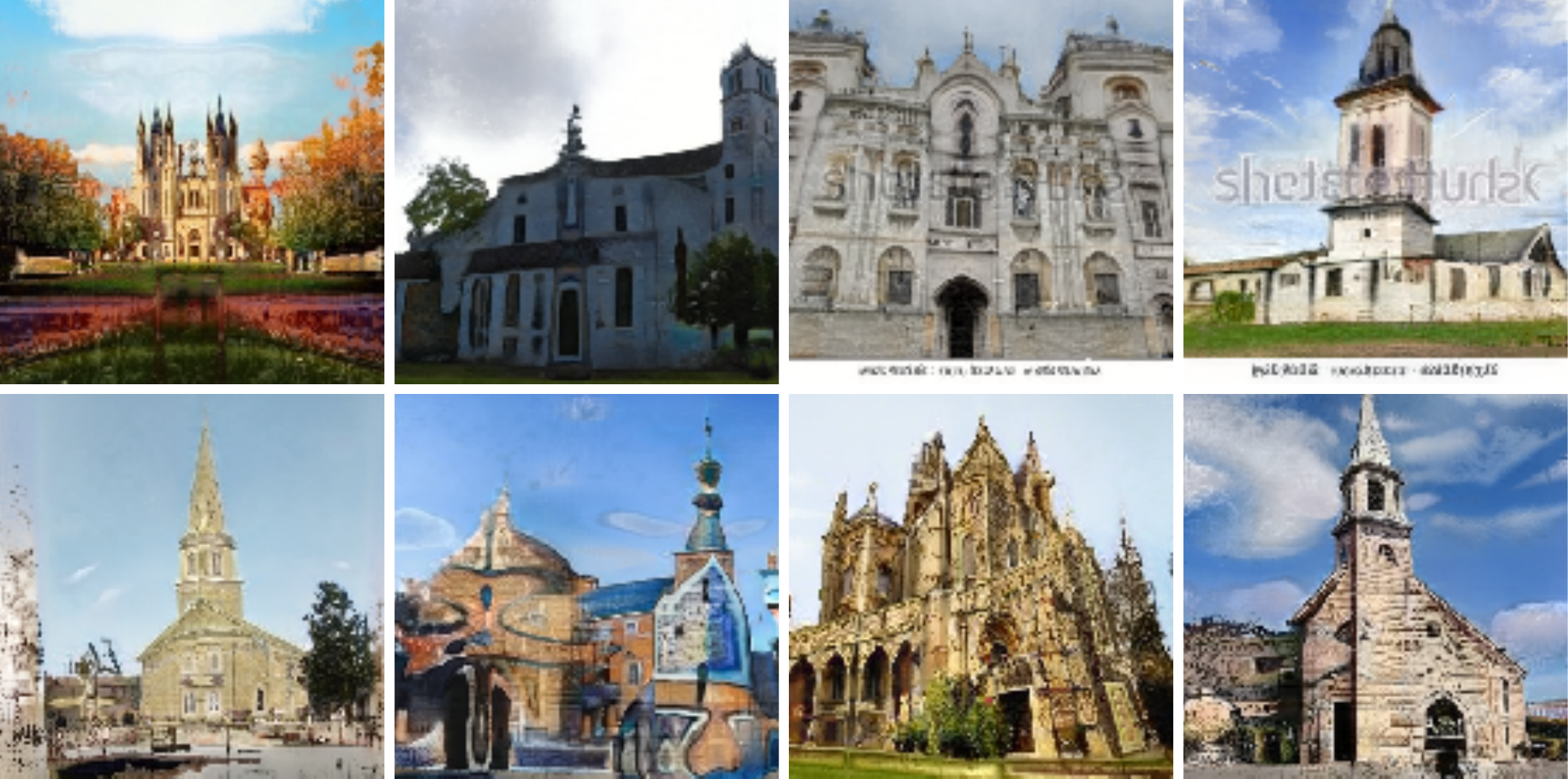} \\
        LSUN-Churches
   \label{fig:patch-church} 
   \vspace{10pt}
\end{subfigure}

\begin{subfigure}[b]{0.45\textwidth}
    \centering
   \includegraphics[width=1\linewidth]{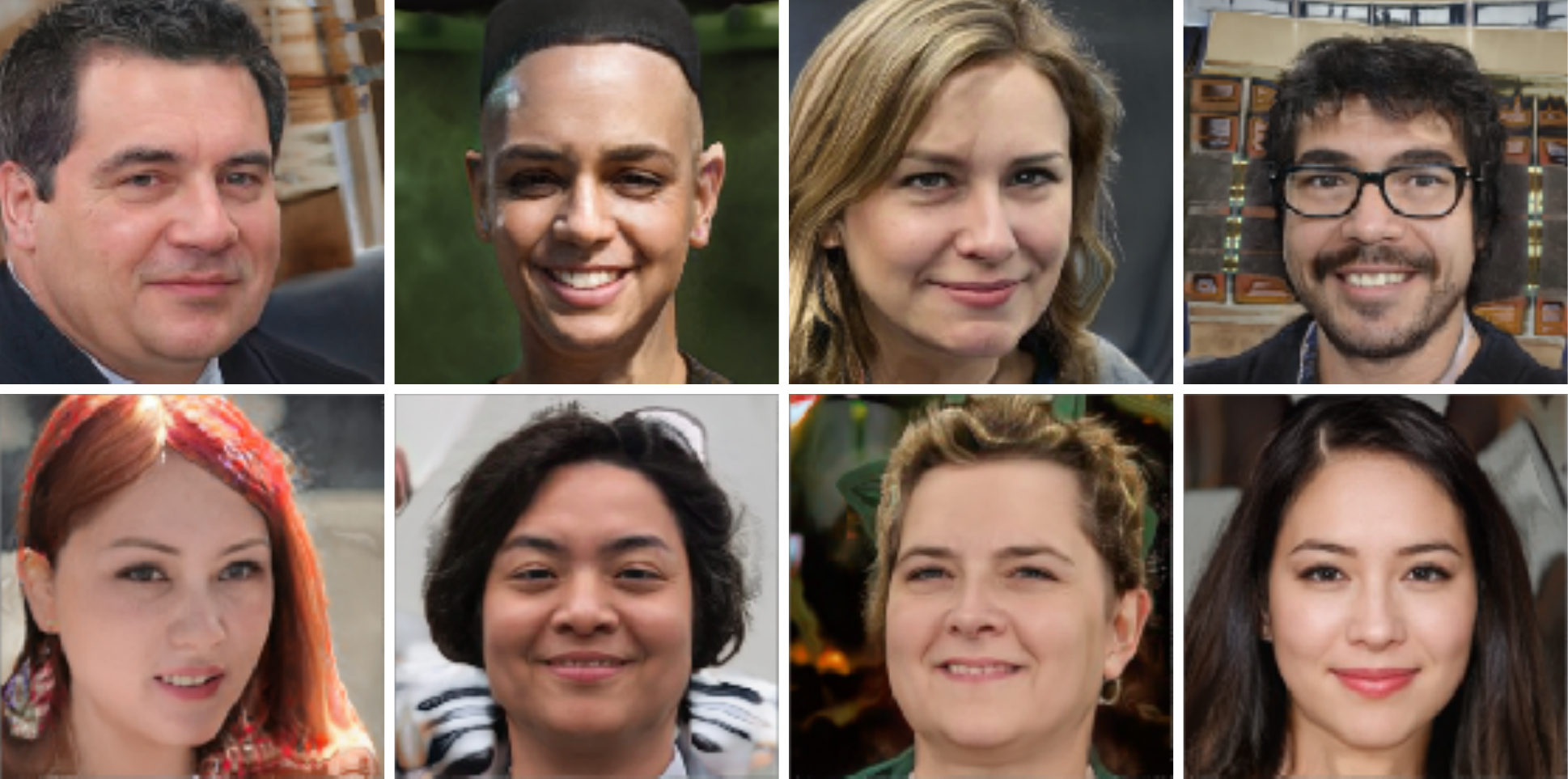} \\
   FFHQ
   \label{fig:patch-ffhq}
\end{subfigure}

    \caption{Samples from CIPS generators learned with memory-constrained \textbf{patch-based training}. Within every grid, the top row contains images from models trained with patches of size $128 \times 128$ and the bottom row represents outputs from training on $64 \times 64$ patches. While the samples obtained with such memory-constrained training are meaningful, their quality and diversity are worse compared to standard training.
    }
    \label{fig:patch_samples}
\end{figure}

Tab.~\ref{tab:patches} reports the quality (FID) for CIPS generators trained on patches of sizes $64\times{}64$ and $128\times{}128$, while the resolution of full images equals $256\times{}256$. 
Fig.~\ref{fig:patch_samples} shows the outputs of models, trained with the patch-based pipeline. In our experiments, training with smaller size of patches degrades the overall quality of resulting samples.

\section{Additional samples}
\label{sec:samples}

\begin{figure*}[ht] 
  \begin{minipage}[b]{0.45\linewidth}
    \centering
    \includegraphics[width=\linewidth]{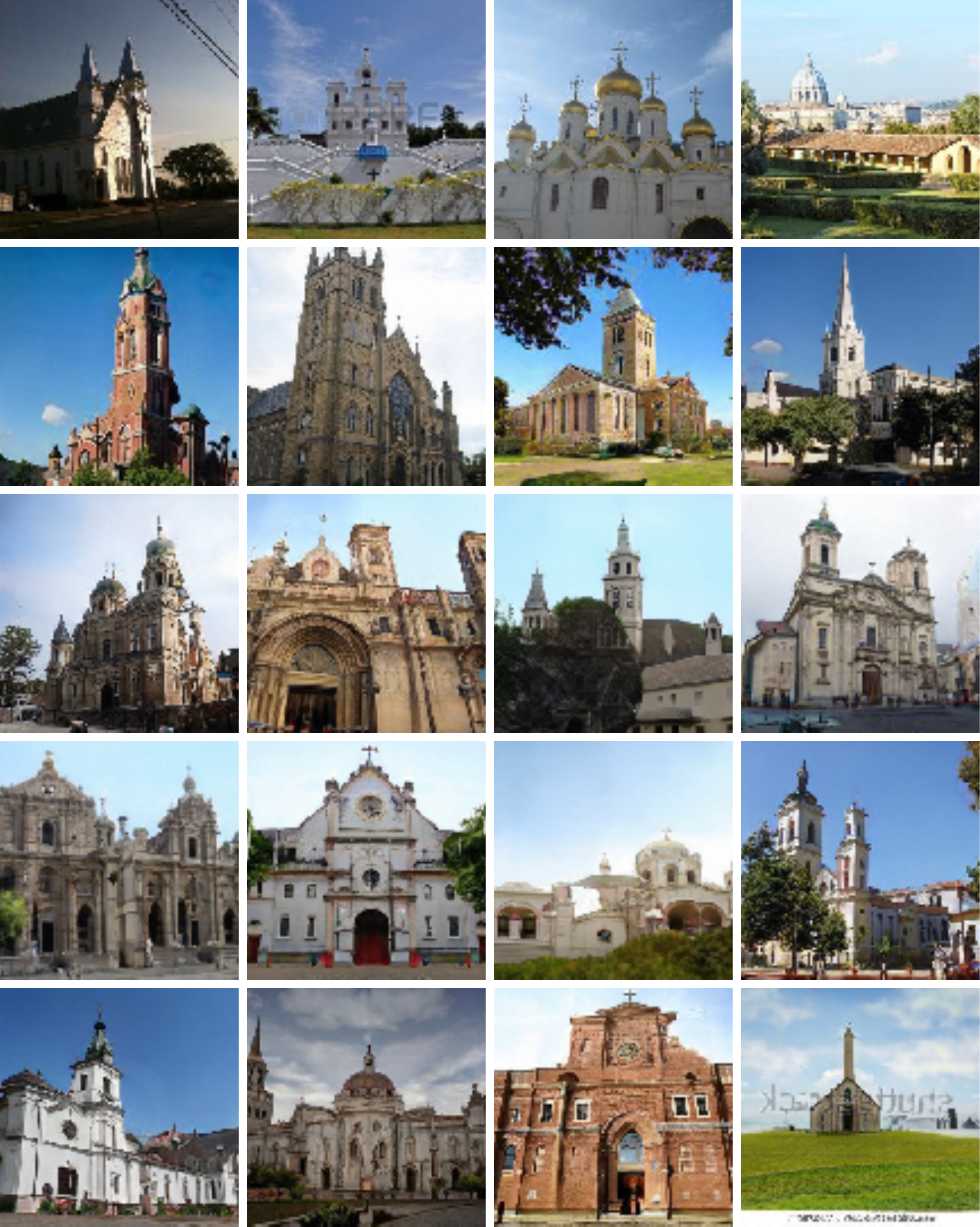} \\
    LSUN-Churches
    \label{fig:samples_churches} 
    \vspace{4ex}
  \end{minipage}
  \hfill
  \begin{minipage}[b]{0.45\linewidth}
    \centering
    \includegraphics[width=\linewidth]{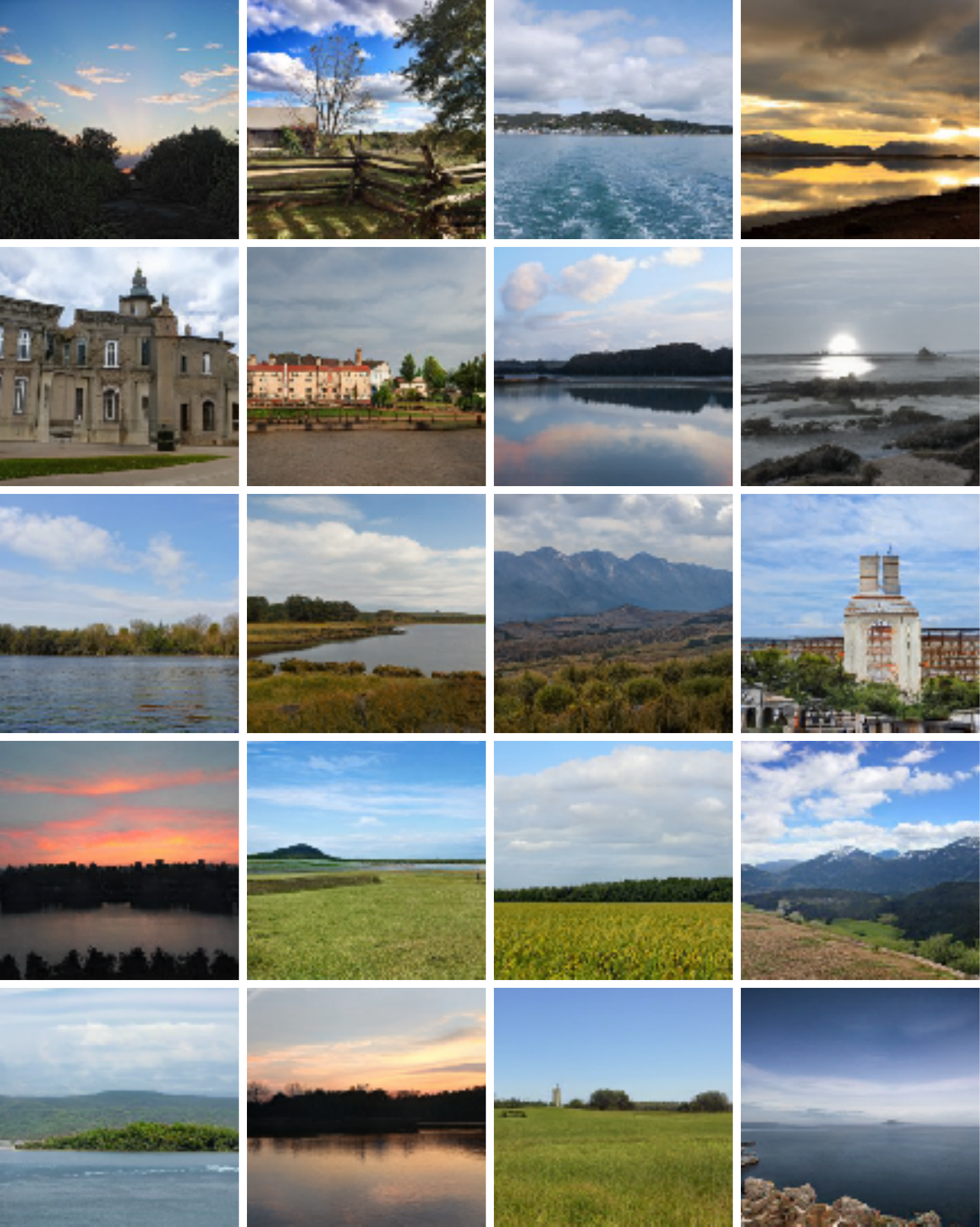} \\
    Landscapes 
    \label{fig:samples_landscapes} 
    \vspace{4ex}
  \end{minipage} 
  
   \begin{minipage}[b]{0.45\linewidth}
    \centering
    \includegraphics[width=\linewidth]{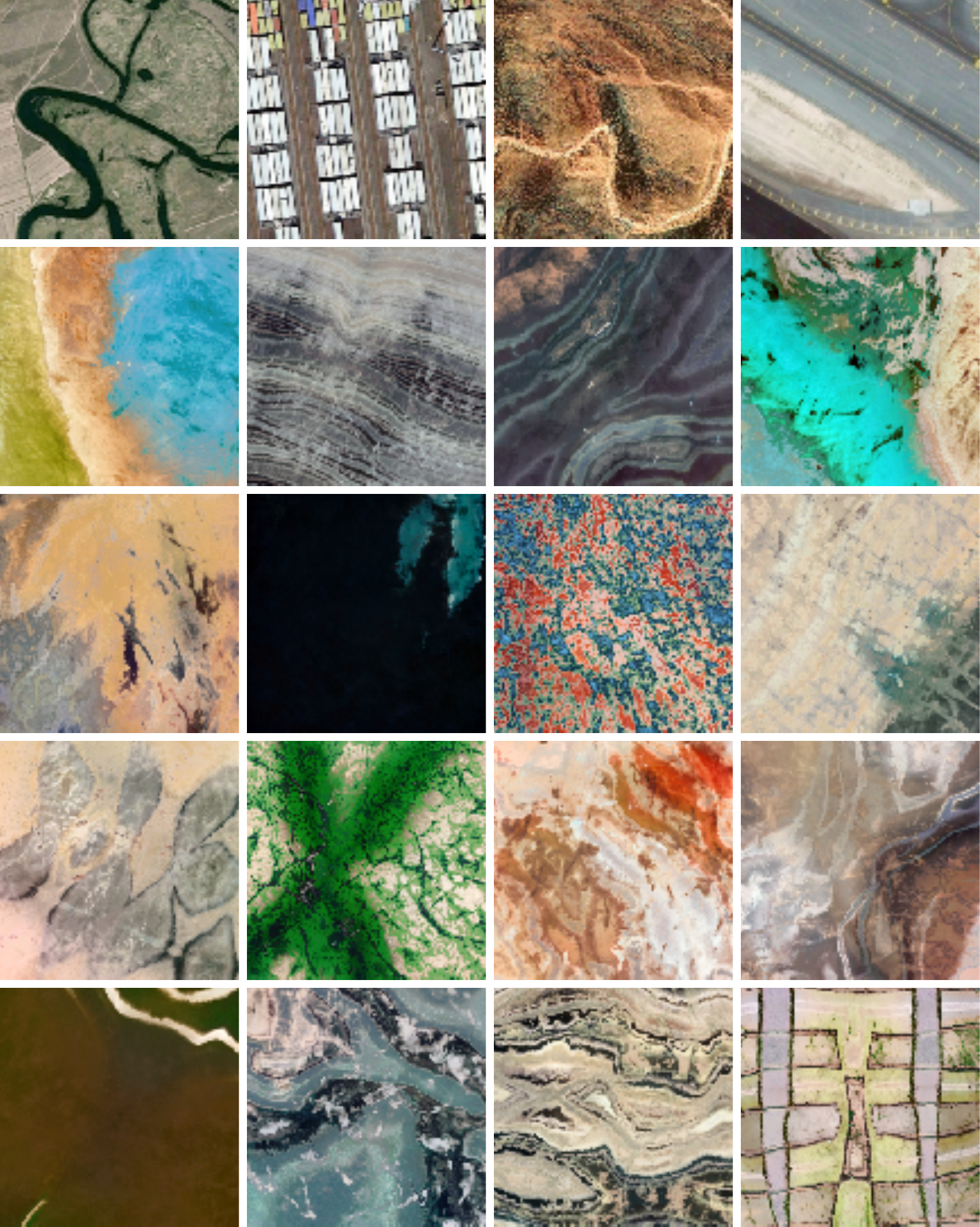} \\
    Satellite-Landscapes
    \label{fig:samples_satellite-landscapes} 
    \vspace{4ex}
  \end{minipage}
  \hfill
  \begin{minipage}[b]{0.45\linewidth}
    \centering
    \includegraphics[width=\linewidth]{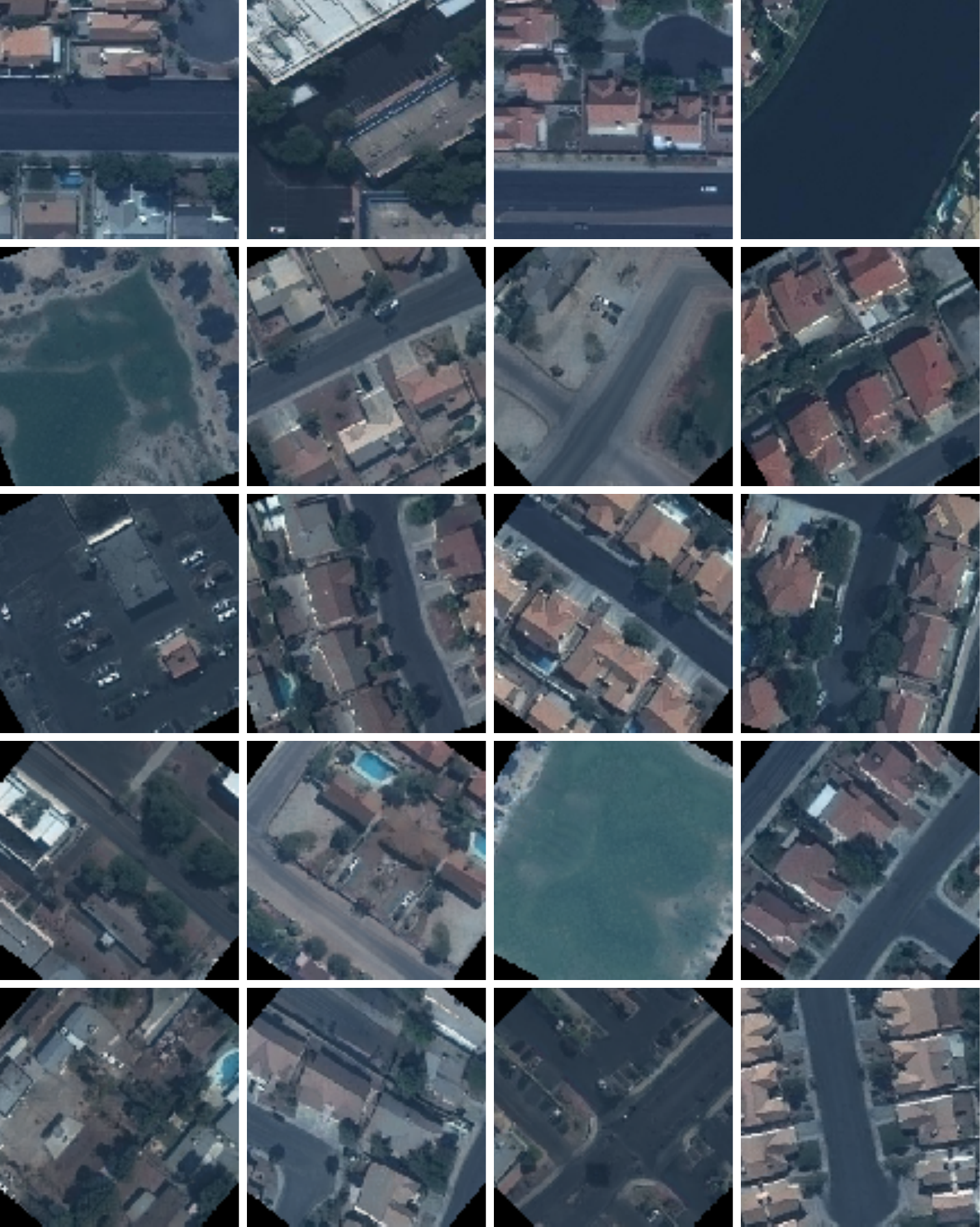}\\
    Satellite-Buildings
    \label{fig:samples_satellite-buildings} 
    \vspace{4ex}
  \end{minipage} 
 \caption{Samples from CIPS generators trained on various datasets. The top row of every grid shows real samples, and the remaining rows contain samples from the models. The samples from CIPS generators are plausible and diverse.}
 \label{fig:samples}
\end{figure*}

\begin{figure*}
    \centering
    \includegraphics[width=0.85\linewidth]{supplementary_figures/panorama_blends/panorama_25_2.png} \\
    \includegraphics[width=0.85\linewidth]{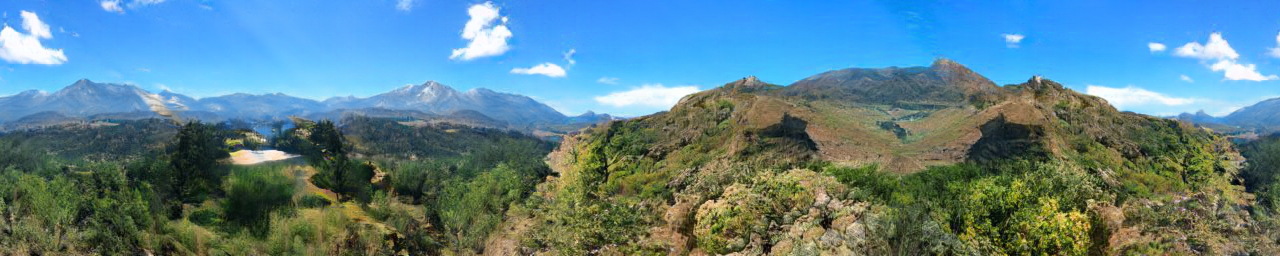} \\
    \includegraphics[width=0.85\linewidth]{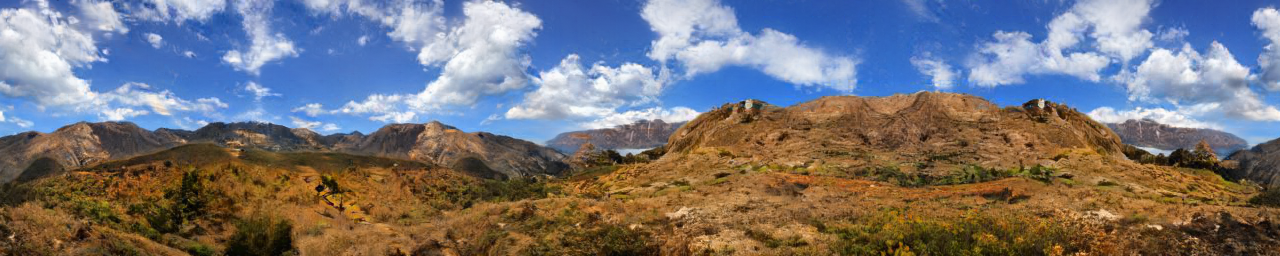}
    \caption{Additional samples of cylindrical panoramas, generated by the CIPS model trained on the Landscapes dataset. The training data contains standard landscape photographs from the Flickr website. No panoramas are provided to the model during training. }
    \label{fig:panoramas}
\end{figure*}

In Fig.~\ref{fig:samples}, we provide additional samples from CIPS generators trained on different datasets. We also demonstrate more samples of cylindrical panoramas in Fig.~\ref{fig:panoramas}.

\begin{figure*}
    \centering
    \includegraphics[width=0.85\linewidth]{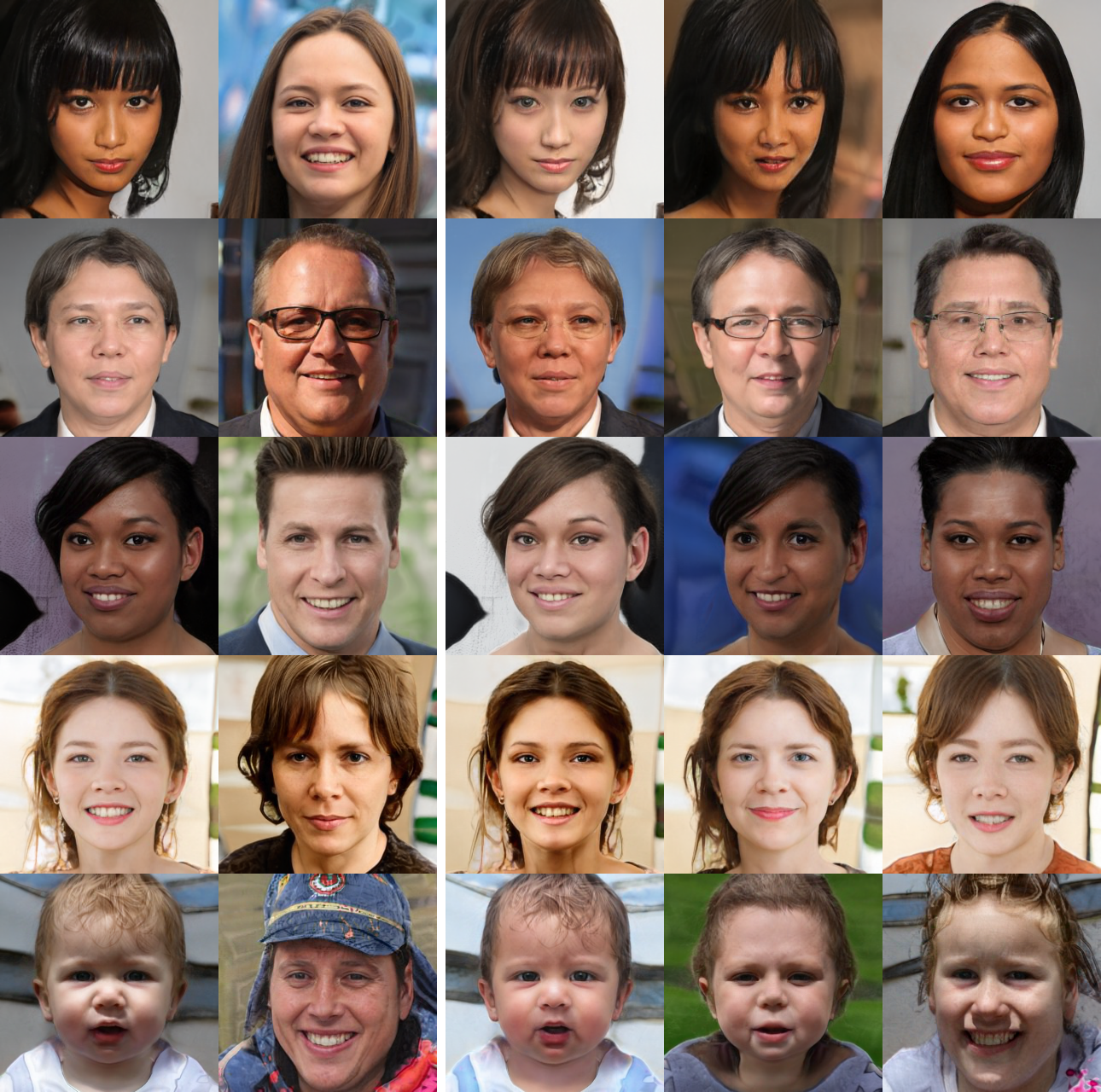}
    \caption{\textbf{Layer-wise style mixing}. The two leftmost columns contain source images A and B. In the rightmost three columns, we replace the latent code $\mathbf{w}$ of A with the latent code $\mathbf{w}$ of B at layers (left to right): 6-8, 3-5, 1-2. The visualization suggests that layers 1-2 control the pose and the shape of the head, the middle layers (3-5) control finer geometry such as the shape of eyes, eyebrows and nose, and the final layers (6-8) controls the skin color and the textures. Interestingly, this CIPS model was trained without layerwise mixing, and therefore such decomposition likely arises from the architectural prior.}
    \label{fig:style_mixing}
\end{figure*}

Although we do not apply mixing regularization~\cite{stylegan1} at train time, our model is still capable of layer-wise combination of latent variables at various depth (see Fig.~\ref{fig:style_mixing}). The examples suggest that similarly to StyleGAN, different layers of CIPS control different aspects of images.
\section{Nearest neighbors}
\label{sec:neighbors}

\begin{figure*}
    \centering
    \makeatletter
    \@for\imagenumber:={1,3,4,5}\do{
        \begin{subfigure}[h]{0.85\linewidth}
            \centering
            \includegraphics[width=\linewidth]{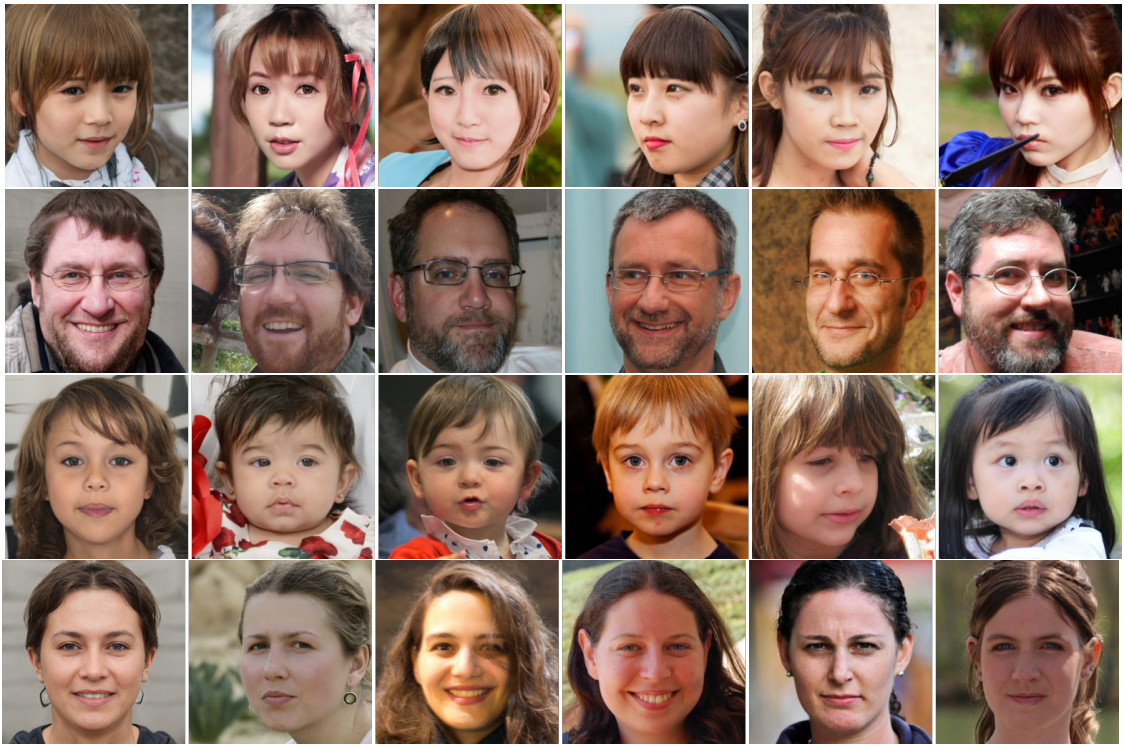}
        \end{subfigure}
    }
    \makeatother
    
    \caption{\textbf{Nearest neighbors for generated faces}. 
        Within each row, we show a sample from the model on the left. The remaining columns contain real images that are closest to the respective sample in terms of the FaceNet~\cite{facenet} descriptor. The visualization suggests that the CIPS model generalizes well beyond memorization of the training dataset.
    }
    \label{fig:neighbors}
\end{figure*}

To assess the generalization ability of CIPS architecture, we also show the samples from the model trained on the FFHQ face dataset alongside the most similar faces from the train dataset. To mine the most similar faces, we extract faces using the MTCNN model~\cite{mtcnn}, and then compute their embeddings using FaceNet~\cite{facenet} (the public implementation of these models\footnote{\url{https://github.com/timesler/facenet-pytorch}} was used).
Fig.~\ref{fig:neighbors} shows five nearest neighbors (w.r.t.\ FaceNet descriptors) for each samples. The samples generated by the model are clearly not duplicates of the training images.

\end{appendices}
\end{document}